\documentclass{sig-alternate-05-2015}
\usepackage{hyperref}
\usepackage{graphicx}
\usepackage{caption}
\usepackage{subcaption}

\begin{document}
\setcopyright{acmcopyright}
\doi{}
\conferenceinfo{SuCCESS'16}{September 12, 2016, Leipzig, Germany}
\CopyrightYear{2016}
\title{A Physical Metaphor to Study Semantic Drift}
\numberofauthors{5}
\author{
\alignauthor S\'andor Dar\'anyi\\
       \affaddr{Swedish School of Library and Information Science}\\
			 \affaddr{University of Bor\aa s}\\
       \affaddr{Bor\aa s 50190, Sweden}\\
       \email{sandor.daranyi@hb.se}
\alignauthor Peter Wittek\\
       \affaddr{Swedish School of Library and Information Science}\\
			 \affaddr{University of Bor\aa s}\\
       \affaddr{Bor\aa s 50190, Sweden}\\
       \affaddr{ICFO-The Institute of Photonic Sciences}\\
       \affaddr{Castelldefels 08860, Spain}
\alignauthor Konstantinos Konstantinidis\\
       \affaddr{Information Technologies Institute}\\
       \affaddr{The Centre for Research \& Technology, Hellas}\\
       \affaddr{Thessaloniki 57001, Greece}\\
       \email{konkonst@iti.gr}
\and 
\alignauthor Symeon Papadopoulos\\
       \affaddr{Information Technologies Institute}\\
       \affaddr{The Centre for Research \& Technology, Hellas}\\
       \affaddr{Thessaloniki 57001, Greece}\\
       \email{papadop@iti.gr}
\alignauthor Efstratios Kontopoulos\\
       \affaddr{Information Technologies Institute}\\
       \affaddr{The Centre for Research \& Technology, Hellas}\\
       \affaddr{Thessaloniki 57001, Greece}\\
       \email{skontopo@iti.gr}
}
\date{}

\maketitle
\begin{abstract}
In accessibility tests for digital preservation, over time we experience drifts of localized and labelled content in statistical models of evolving semantics represented as a vector field. This articulates the need to detect, measure, interpret and model outcomes of knowledge dynamics. To this end we employ a high-performance machine learning algorithm for the training of extremely large emergent self-organizing maps for exploratory data analysis. The working hypothesis we present here is that the dynamics of semantic drifts can be modeled on a relaxed version of Newtonian mechanics called social mechanics. By using term distances as a measure of semantic relatedness vs. their PageRank values indicating social importance and applied as variable `term mass', gravitation as a metaphor to express changes in the semantic content of a vector field lends a new perspective for experimentation. From `term gravitation' over time, one can compute its generating potential whose fluctuations manifest modifications in pairwise term similarity vs. social importance, thereby updating Osgood's semantic differential. The dataset examined is the public catalog metadata of Tate Galleries, London.
\end{abstract}



\section{Introduction}
The evolving nature of digital collections comes with an extra difficulty: due to various but constant influences inherent in updates, the interpretability of the data keeps on changing. This manifests itself as concept drift~\cite{wang2011concept} or semantic drift~\cite{wittek2015monitoring, gulla2010semantic}, the gradual change of a concept's semantic value as it is perceived by a community. Despite terminology differences, the problem is real and with the increasing scale of digital collections, its importance is expected to grow~\cite{schlieder2010digital}. If we add drifts in cultural values as well, the fallout from their combination brings memory institutions in a vulnerable position as regards long term digital preservation. We illustrate this on a museum example, the subject index of the Tate Galleries, London.
In our example, semantic drifts lead to limited access by Information Retrieval (IR). The methodology we apply to demonstrate our point is vector field semantics by emergent self-organizing maps (ESOM)~\cite{ultsch2005clustering}, because the interpretation of semantic drift needs a theory of update semantics~\cite{veltman1996defaults}, integrated with a vector field rather than a vector space representation of content~\cite{wittek2014vector,wittek2015monitoring}. Further, given such content dynamics, we argue that for its modeling, one can fall back on tested concepts from classical (Newtonian) mechanics and differential geometry. For such a framework, e.g. similarity between objects or features can be considered an attractive force, and changes over time manifest in content drifts have a quasi-physical explanation.
The main contributions of this paper are the following:
\begin{enumerate}
	\item A methodology for the detection, measurement and interpretation of semantic drift;
	\item On drift examples, an improved understanding of how semantic content as a vector field `behaves' over time by falling back on physics as a metaphor;
	\item As a consequence of the above, the concept of semantic potential as a combined measure of semantic relatedness and semantic importance.
\end{enumerate}

\section{Background}
\subsection{Terminology}
Evolving semantics (also often referred to as `semantic change'~\cite{tury2006approach}) is an active and growing area of research into language change~\cite{baker08languageChange} that observes and measures the phenomenon of changes in the meaning of concepts within knowledge representation models, along with their potential replacement by other meanings over time. Therefore it can have drastic consequences on the use of knowledge representation models in applications. Semantic change relates to various lines of research such as ontology change, evolution, management and versioning~\cite{merono2013detecting}, but it also entails ambiguous terms of slightly different meanings, interchanging shifts with drifts and versioning, and applied to concepts, semantics and topics, always related to the thematic composition of collections~\cite{yildiz2006ontology,uschold2000creating,klein2001ontology}. A related term is semantic decay as a metric: it has been empirically shown that the more a concept is reused, the less semantically rich it becomes~\cite{pareti2015linked}. Though largely counter-intuitive, this derivation is based on the fact that frequent usage of terms in diverse domains leads to relaxing the initially strict semantics related to them. The opposite would hold if a term was persistently used within a single domain (or in to a great extent similar domains), which would lead to its gradual specialization and enrichment of its semantics. 

\subsection{Related Research}
Here we mention four relevant directions, all of them contributors to our understanding of a complex issue in their overlap.

\subsubsection{Temporality and Advanced Access}
By advanced access to digital collections we mean the spectrum of automatic indexing, automatic classification, IR, and information visualization. All of the aforementioned can have a temporal aspect: trend analysis, emergence of concepts or ideas, representation of the past and the future, network dynamic, shaping and decay of communities, and in general, any Web research topic where a dynamic understanding is superior to a static view, requires integration of the time dimension. Examples comprise e.g the presentation, organization and exploration of search results~\cite{alonso2011temporal} in the context of web dynamics and analytics including the dynamics of user behaviour~\cite{radinsky2013temporal}; interacting with ephemeral content of the historical web~\cite{adar2008zoetrope}, visualizing the evolution of image content tags~\cite{dubinko2007visualizing}, or temporal topic detection without citation analysis~\cite{shaparenko2005identifying}. A related but separate research area for the above is in the overlap of cultural heritage and IR~\cite{koolen2009information, jong2005temporal}.

\subsubsection{Vector Space vs. Vector Field Semantics}
For an IR model to be successful, its relationship with at least one major theory of word meaning has to be demonstrated. With no such connection, meaning in numbers becomes the puzzle of the ghost in the machine. For the vector space IR model (VSM) - underlying many of today's competitive IR products and services - such a connection can be demonstrated; for others like PageRank~\cite{brin1998pagerank}, the link between graph theory and linear algebra leads to the same interpretation. Namely, in both cases, the theory of word semantics cross-pollinating numbers with meaning is of a contextual kind, formalized by the distributional hypothesis~\cite{harris1968msl} which posits that words occurring in similar contexts tend to have similar meanings. As a result, the respective models can imitate the field-like continuity of conceptual content. However, unless we consider the VSM roots of both the probabilistic relevance model\footnote{Because it departs from a `binary index descriptions of documents', see~\cite{robertson1976relevance}.} and its spinoffs including BM25,\footnote{See p. 339 in~\cite{robertson2009probabilistic}.} such a link is still waiting to be shown between probability and semantics~\cite{frommholz2010supporting}. 

Although several attempts exist to this end~\cite{turney2010frequency, pulman2013semantic}, a brief overview should be helpful. Looking for a good fit with some reasonably formalized theory of semantics, two immediate questions emerge. First, can the observed features be regarded as entries in a vocabulary? If so, distributional semantics applies and, given more complex representations, other types may do so as well~\cite{wittek2013complex}. The second question is, do they form sentences? For example, one could regard a workflow (process) a sentence, in which case compositional semantics applies~\cite{coecke2010mathematical, sadrzadeh2011compositional}. If not, theories of word semantics should be considered only. Below we shall depart from this assumption.

Notwithstanding the fact that vector space in its most basic form is not semantic, its ability to yield results which make sense goes back to the fact that the context of sentence content is partially preserved even after having eliminated stop-words which are useless for document indexing. This means that Wittgenstein's contextual theory of meaning (`Meaning is use') holds~\cite{wittgenstein1967pi}, also pronounced by the distributional hypothesis. This is exploited by more advanced vector based indexing and retrieval models such as Latent Semantic Analysis (LSA)~\cite{deerwester1990indexing} or random indexing~\cite{kanerva2000random}, as well as by neural language models, ranging from the Simple Recurrent Networks, and their very popular flavour, Long Short-Term Memory~\cite{hochreiter1997long}, or the recently proposed Global Vector for Word Representation~\cite{pennington2014glove}, which are currently considered to be the state-of-the-art approach for text representation. However, we should also remember another approach paraphrased as `Meaning is change', namely the stimulus-response theory of meaning proposed e.g. by Bloomfield\footnote{\href{http://en.wikipedia.org/wiki/Leonard\_Bloomfield}{en.wikipedia.org/wiki/Leonard\_Bloomfield}} in anthropological linguistics and Morris \footnote{\href{http://en.wikipedia.org/wiki/Charles\_W.\_Morris}{en.wikipedia.org/wiki/Charles\_W.\_Morris}} in behavioral semiotics, plus the biological theory of meaning~\cite{uexkull2013streifzuge}. These authors stress that the meaning of an action is in its consequences. Consequently word semantics should be represented not as a vector space with position vectors only, but as a dynamic vector field with both position and direction vectors~\cite{wittek2014vector}.

\subsubsection{Linguistic `Forces'}
As White suggests, linguistics, like physics, has four binding forces~\cite{white2002cross}:
\begin{enumerate}
	\item The strong nuclear force, which is the strongest `glue' in physics, corresponds to word uninterruptability (binding morphemes into words);
	\item Electromagnetism, which is less strong, corresponds to grammar and binds words into sentences;
	\item The weak nuclear force, being even less strong, compares to texture or cohesion (also called coherence), binding sentences into texts;
	\item Finally gravity as the weakest force acts like intercohesion or intercoherence which binds texts into literatures (i.e. documents into collections or databases).
\end{enumerate}

Mainstream linguistics traditionally deals with Forces 1 and 2, while discourse analysis and text linguistics are particularly concerned with Force 3. The field most identified with the study of Force 4 is information science. As the concept of force implies, referring here to attraction, it takes energy to keep things together, therefore the energy doing so is stored in agglomerations of observables of different kinds in different magnitudes, and can be released from such structures. A notable difference between physical and linguistic systems is that extracting work content, i.e. `energy' from symbols by reading or copying them does not annihilate symbolic content.
Looking now at the same problem from another angle, in the above and related efforts, `energy' inherent in all four types can be the model of e.g. a Type 2, i.e. electromagnetism-like attractive-repulsive binding force such as lexical attraction, also known as syntactic word affinity~\cite{beeferman1997mod} or sentence cohesion, such as by modeling dependency grammar by mutual information~\cite{yuret1998discovery}. In a text categorization and/or IR setting, a similar phenomenon is term dependence based on their co-occurrence. 

\subsubsection{Semantic Kernels and `Gravity'}
A radial basis function (RBF) kernel, being an exponentially decaying feature transformation, has the capacity to generate a potential surface and hence create the impression of gravity, providing one with distance-based decay of interaction strength, plus a scalar scaling factor for the interaction, i.e. \begin{math}K(x,x')=exp(-γ||x-x'||^{2})\end{math}~\cite{moschitti2010kernel}. We know that semantic kernels and the metric tensor are related, hence some kind of a functional equivalent of gravitation shapes the curvature of classification space~\cite{amari1999improving, eklund2016context}. At the same time, gravitation as a classification paradigm~\cite{peng2009data} or a clustering principle~\cite{aghajanyan2015gravitational} is considered as a model for certain symptoms of content behavior.

\section{WORKING HYPOTHESIS \& METHODOLOGY}
In order to combine semantics from computational linguistics with evolution, we select the theory of semantic fields~\cite{trier1934sprachliche} and blend it with multivariate statistics plus the concept of fields in classical mechanics to bring it closer to Veltman's update semantics~\cite{veltman1996defaults}, and to enable machine learning. Our working hypothesis for experiment design is as follows:
\begin{itemize}
	\item Semantic drifts can be modeled on an evolving vector field as suggested by~\cite{wittek2015monitoring, wittek2014vector};
	\item To follow up on the analogy from semantic kernels defining the curvature of classification space and let this curvature evolve, Newton's universal law of gravitation can be adapted to the idea of the dynamic library~\cite{salton1975dynamic}. To this end, we model similarity by \begin{math}F = Gm_{1}m_{2}/r^{2}\end{math}, with term dislocations over epochs stored in distance matrices. Ignoring G, we shall use the PageRank value of index terms on their respective hierarchical levels for mass values. Since force is the negative gradient of potential, i.e. \begin{math}F\left(x\right)=-dU/dx\end{math}, we can compute this potential surface over the respective term sets to conceptualize the driving mechanism of semantic drifts;
	\item The potential following from the gravity model manifests two kinds of interaction between entries in the indexing vocabulary of a collection. Over time, changes in collection composition lead to different proportions of semantic similarity vs. authenticity between term pairs, expressed as a cohesive force between features and/or objects.
\end{itemize}

\subsection{ESOMs and Somoclu}
\subsubsection{Vector Field Creation by ESOMs}
In the various flavours of the VSM, we work with an $m\times n$ matrix in which columns are indexed by documents and rows by terms. We shall focus here on the $m$ term vectors only, which identify specific locations in the $n$-dimensional space spanned by the documents.

A scalar or vector field is defined at all points in space, so it is insufficient to have a value at the discrete locations identified by the term vectors. To assign a vector value to each point in space, we work on a two-dimensional surface. All term vectors have a location on this surface. All the other points on the surface which do not have a vector assigned to them are interpolated.

The assignment of points on the surface and the term vectors is done by training a self-organizing map, that is, a grid of artificial neurons. Each node in the grid is associated with a weight vector of $n$ dimensions, matching the term vectors. Taking a term vector, we search for the closest weight vector, and pull it slightly closer to the term vector, repeating the procedure with the weight vectors of the neighboring neurons, with decreasing weight as we get further away from the best matching unit. Then we take the next term vector and repeat this from finding the best matching unit until every term vector is processed. We call a training round that uses all term vectors an epoch. We can have subsequent training epochs with a smaller neighborhood radius and a lower learning rate. While there is no criterion for a convergence, we can continue training epochs until the topology of the network no longer shows major changes. The resulting map reflects the local topology of the original high-dimensional space~\cite{kohonen2001som}.

Since we would like to train large maps to get a meaningful approximation in the space between term vectors, we turn to a high-performance implementation called Somoclu\footnote{\url{https://peterwittek.github.io/somoclu/}}\cite{wittek2013somoclu}.

\subsubsection{Drift Detection}
The task of drift detection, measurement and interpretation is carried out in three basic steps as follows:
\begin{itemize}
	\item Step 1: Somoclu maps the high-dimensional topology of multivariate data to a low-dimensional (2-d) embedding by ESOM. The algorithm is initialized by LSA, Principal Component Analysis (PCA), or random indexing, and creates a vector field over a rectangular grid of nodes of an artificial neural network, adding continuity by interpolation among grid nodes. Due to this interpolation, content is mapped onto those nodes of the neural network that represent best matching units (BMUs). 
	\item Step 2: Clustering over this low-dimensional topology marks up the cluster boundaries to which BMUs belong. Their clusters are located within ridges or watersheds~\cite{ultsch2005clustering, tosi2014probability, lotsch2014exploiting}. Content splitting tendencies are indicated by the ridge wall width and height around such basins so that the method yields an overlay of two aligned contour maps in change, i.e. content structure vs. tension structure. In Somoclu, nine clustering methods are available. Because self-organizing maps, including ESOM, reproduce the local but not the global topology of data, the clusters should be locally meaningful and consistent on a neighborhood level only.
	\item Step 3: Evolving cluster interpretation by semantic consistency check can be measured relative to an anchor (non-shifting) term used as the origin of the 2-d coordinate system, or by distance changes from a cluster centroid, etc. In parallel, to support semiautomatic evaluation, variable cluster content can be expressed for comparison by histograms, pie diagrams, or other visualization methods. 
\end{itemize}

\section{DATASET AND EXPERIMENT DESIGN}

\subsection{Tate Subject Index}
Tate holds the national collection of British art from 1500 to the present day and international modern and contemporary art. The collection embraces all media, from painting, drawing, sculpture and prints to photography, video and film, installation and performance. The 19th century holdings are dominated by the Turner Bequest with cca 30,000 works of art on paper, including watercolors, drawings and 300 oil paintings. The catalog metadata for the 69,202 artworks that Tate owns or jointly owns with the National Galleries of Scotland are available in JSON format as open data.\footnote{\href{http://github.com/tategallery/collection}{github.com/tategallery/collection}} Out of the above, 53,698 records are timestamped. The artefacts are indexed by Tate's own hierarchical subject index which has three levels, from general to specific index terms.\footnote{\href{http://www.tate.org.uk/art/artworks/turner-self-portrait-n00458}{tate.org.uk/art/artworks/turner-self-portrait-n00458}}

\subsection{Analysis Framework Description}
To study the robust core of a dynamically changing indexing vocabulary, we filtered the dataset for a start. As statistics for the Tate holdings show two acquisition peaks in 1796-1844 (33,625 artworks) and 1960-2009 (12,756 artworks), we focused on these two periods broken down into 10 five-years epochs each, with altogether 46,381 artworks. In the 19th century period, subject index level 1 had 22 unique general index terms (21 of them persistent over ten epochs), level 2 had 203 unique intermediate index terms (142 of them persistent), and level 3 had 6624 unique specific index terms (225 of them persistent). In the 20th century period, level 1 had 24 unique terms (22 of them persistent), level 2 used 211 unique terms (177 of them persistent), and level 3 had 7536 unique terms (288 of them persistent over ten epochs). Table \ref{Tab:1} displays a sample entry from the subject index. 
Following text pre-processing, which included the application of tokenization and stop-word removal on all three levels of concepts in the subject index, adjacency matrices and subsequently graphs were created using the co-occurrence of the terms in the artworks as undirected, weighted edges. These matrices were then used to extract an importance measure for each term by employing the PageRank algorithm, and to create ESOM maps using the Somoclu implementation. 

For each of the 80 epochs (2 periods x 4 levels x 10 epochs), the ESOM's codebook was first initialized by employing PCA with randomized SVD, which was then used for mapping the high-dimensional co-occurrence data to an ESOM with a toroid topology. The results were represented on the two-dimensional projection of the toroid using different granularities according to the indexing level (20x12 = level 1, 40x24 = level 2, 50x30 = level 3, 60x40 = all levels together). Introducing the least displaced term per indexing level over a period as an anchor against which all term drifts on that level could be measured, we tracked the tension vs. content structure of evolving term semantics and evaluated the resulting term clusters for their semantic consistency.

\begin{table}[]
\centering
\caption{Sample index terms describing a Turner self-portrait}
\label{Tab:1}
\begin{tabular}{|l|l|l|}
\hline
\begin{tabular}[c]{@{}l@{}}level 1 \\ (general)\end{tabular}    & \begin{tabular}[c]{@{}l@{}}level 2 \\ (intermediate)\end{tabular}        & \begin{tabular}[c]{@{}l@{}}level 3 \\ (specific)\end{tabular} \\ \hline
Objects                                                         & \begin{tabular}[c]{@{}l@{}}Clothing and\\ personal effects\end{tabular}  & Cravat                                                        \\ \hline
People                                                          & Adults                                                                   & Man                                                           \\ \hline
Named individuals                                               & \begin{tabular}[c]{@{}l@{}}Turner, Joseph\\ Mallord William\end{tabular} & -                                                             \\ \hline
Portraits                                                       & Self-portraits                                                           & -                                                             \\ \hline
\begin{tabular}[c]{@{}l@{}}Work and \\ occupations\end{tabular} & \begin{tabular}[c]{@{}l@{}}Arts and \\ entertainment\end{tabular}        & Artist, painter                                               \\ \hline
\end{tabular}
\end{table}

The input matrices were processed by Somoclu as described above and the codebook of each ESOM was clustered using the affinity propagation algorithm. The results were tested for robustness by hierarchical cluster analysis (HCA), using Euclidean distance as similarity measure and farthest neighbor (complete) linkage to maximize distance between clusters, keeping them thereby both distinct and coherent. The ESOM-based cluster maps expressed the evolving semantics of the collection as a series of 2-dimensional landscapes over 10 epochs times two periods.

Term drift detection, measurement and interpretation were based on these maps. To enable drift measurement, we generated a parallel set of maps with the term of greatest importance over all periods as its anchor point. Importance was defined by the Reciprocal Rank Fusion coefficient~\cite{cormack2009reciprocal} which combined the PageRank values of each term over all periods. This relative location was used for the computation of respective term-term distance matrices over every epoch of each period. Term dislocations over epochs were logged, recording both the splits of term clusters mapped onto a single grid node in a previous epoch, or the merger of two formally independent nodes labelled with different terms into a single one. These splits and merges were used to define the drift rate and subsequently the stability of the lexical field.

Finally, as per the second point of the working hypothesis, the gravity and potential surfaces for every epoch were computed. When computing gravity and potential, the property of mass was expressed via each term's PageRank score and the distance by measuring the normalized (sum to 1) Euclidean distance between the corresponding BMU vectors.

\section{RESULTS}
Index term drift detection, measurement and evaluation were based on the analysis of ESOM maps, leading to drift logs on all indexing levels. Parallel to that, covering every time step of collection development, we also extracted normalized histograms to describe the evolving topical composition of the collection, and respective pie charts to describe the thematic composition of the clusters. Further, to check cluster robustness, HCA dendrograms were computed for term-term matrices, also compared with those from term-document matrices. On one hand, these gave us a detailed overview of semantic drift in the analyzed periods. On the other hand, the observed dynamics could be modeled on the gravitational force and its generating potential.

A more detailed report would go beyond the opportunities of this paper. However, some key indications were the following.

\subsection{Semantic Drifts}
Content mapping means that term membership for every cluster in every time step is recorded and term positions and dislocations over time with regard to an anchor position are computed, thereby recording the evolving distance structure of indexing terminology. This amounts to drift detection and its exact measurement. Adding a drift log results in extracted lists of index terms on all indexing hierarchy levels plus their percentage contrasted with the totals. Drifts can be partitioned into splits and merges. In case of a split, two concept labels that used to be mapped on the same grid node in one epoch become separated and tag two nodes in the next phase, while for a merge, the opposite holds. From an IR perspective splits decrease recall and merges decrease precision, limiting the quality of access; from the perspective of long term digital preservation, they indicate at-risk indexing terminology.
Splits and merges were listed by Somoclu for every epoch over both periods. For instance a sample semantic drift log file recorded that due to new entries in the catalog in 1796-1800, by 1800 on subject index level 2, for drifting words i.e. `art', `works', `scientific', `measuring', `monuments', `places', `workspaces'. Therefore, based on the same subject index terms, anyone using this tool in 1800 would have been unable to retrieve the same objects as in 1796.
In a vector field, all the terms and their respective semantic tags are in constant flux due to external social pressures, such as e.g. new topics over items in the collection due to the composition of donations or fashion. Without data about these pressures quasi embedding and shaping the Tate collection, the correlations between social factors and semantic composition of the collection could not be explicitly computed and named. Still, some trends could be visually recognized over both series of maps, going back to their relatively constant semantic structure where temporary content dislocations did not seriously disturb the relationships between terms, i.e. neighboring labels tended to stick with one another, such as `towns, cities, villages' vs. `inland' and `natural'. In other words, the lexical fields as locally represented by Somoclu remained relatively stable.

\begin{figure*}
\centering
        \begin{subfigure}{0.48\textwidth}
                \includegraphics[width=\linewidth]{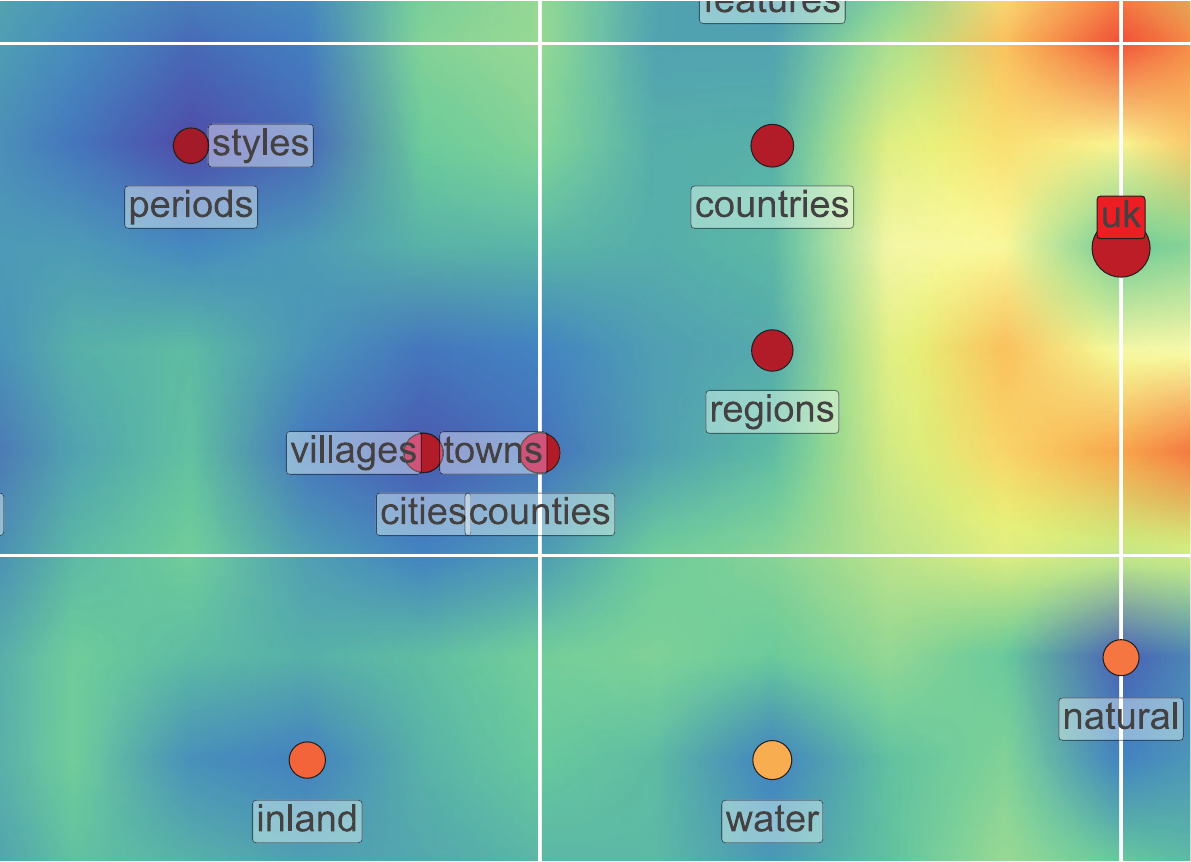}
                \caption{1796-1800}
        \end{subfigure}%
				\hfill
        \begin{subfigure}{0.48\textwidth}
                \includegraphics[width=\linewidth]{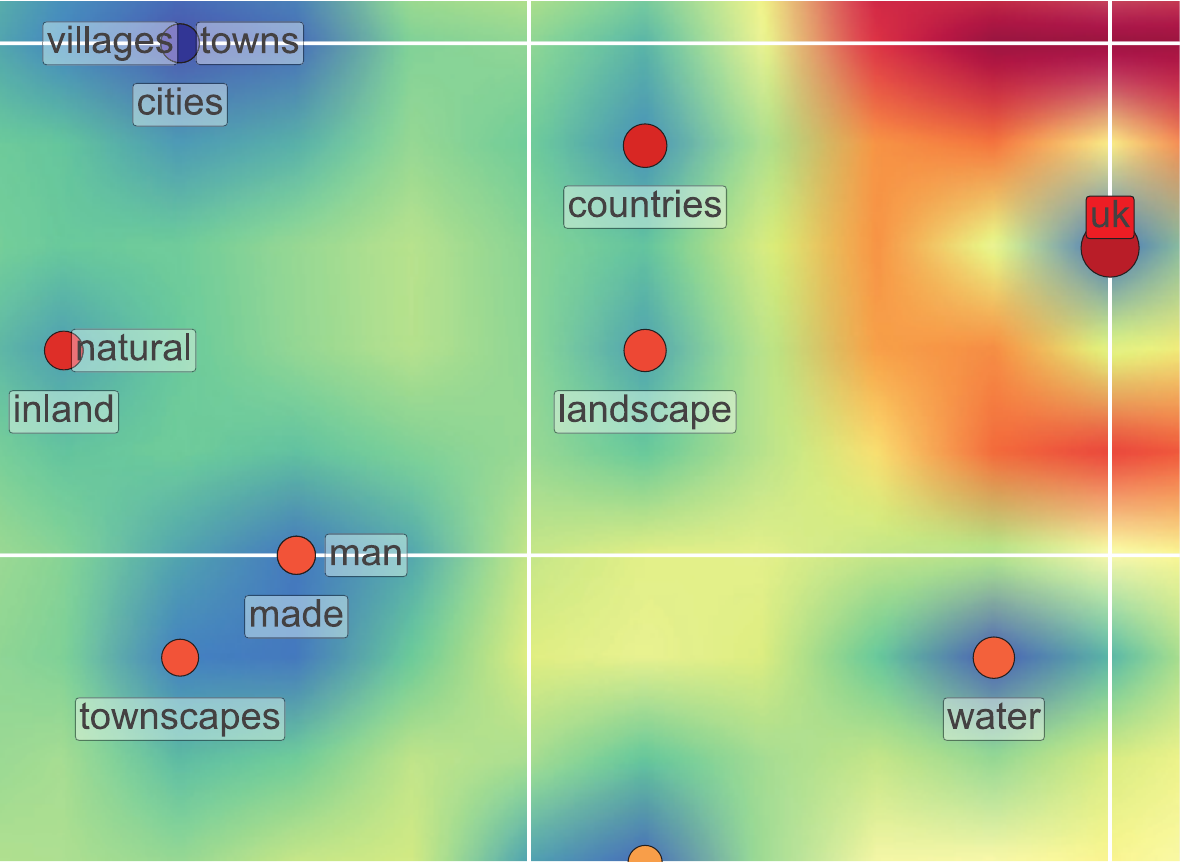}
                \caption{1801-1805}
        \end{subfigure}%
        \caption{Excerpt from the tension vs. content structure changes in the level 2 (intermediate) index term landscape in 1796-1805. Blue basins host content, brown ridges indicate tensions. Whereas `towns', `cities', `villages' remain merged over both epochs, `inland' and `natural' become merged by 1805.}
				\label{Fig:1}
\end{figure*}

The stability of these fields was measured in terms of drift rates which were computed by detecting the splits and merges that happened to the BMUs (e.g. \ref{Fig:1}). Specifically, we were not looking at the distance they travelled, rather at the fact that they formed or joined or moved away from a cluster (i.e. a BMU) in between epochs.

Overall, in this particular collection, splits between level 1 concepts took place occasionally, whereas both splits and merges occurred on indexing levels 2-3 on a regular basis. The drift rate was increasingly high: for level 2 index terms, it was 19-22~\% in the 1796-1845 period vs. 15-27.5~\% in 1960-2009, whereas for level 3 terms it was 29-57~\% (1796-1845) vs. 54-61~\% (1960-2009). These percentages suggest that the more specific the subject index becomes, the more volatile its terminology, especially with regard to modern art.

\subsection{Content vs. Tension Structure}
To describe the composition of the social tensions shaping this collection, one can compare e.g. the level 2 indexing vocabularies for both periods. In general, this is where one witnesses the workings of language change, part producing new concepts, part letting certain index terms decay. E.g. focus is shifting from a concept to its variant (e.g. `nation' to `nationality'), a renaissance of interest in the transcendent beyond traditional notions of religion and the supernatural (`occultism', `magic', `tales'), fascination for the new instead of the old, or a loss of interest in `royalty' and `rank'. Toys and concepts like `tradition', the `world', `culture', `education', `films', `games', `electricity' and `appliances' make a debut in art. A representation of such tendencies in content change with manifest tensions is visualized in Figure \ref{Fig:1}. Here, tendency means a projected possible, but not necessarily continuous, trend - should the composition of the collection continue to evolve over the next epoch like it used to develop over the past one, the indicated splits and merges would be more probable to form new content agglomerations than random ones. 

\subsection{Content Dynamics}
\begin{figure*}
\centering
(a)\\
        \begin{subfigure}[b]{0.20\textwidth}
                \centering
                \includegraphics[width=\linewidth]{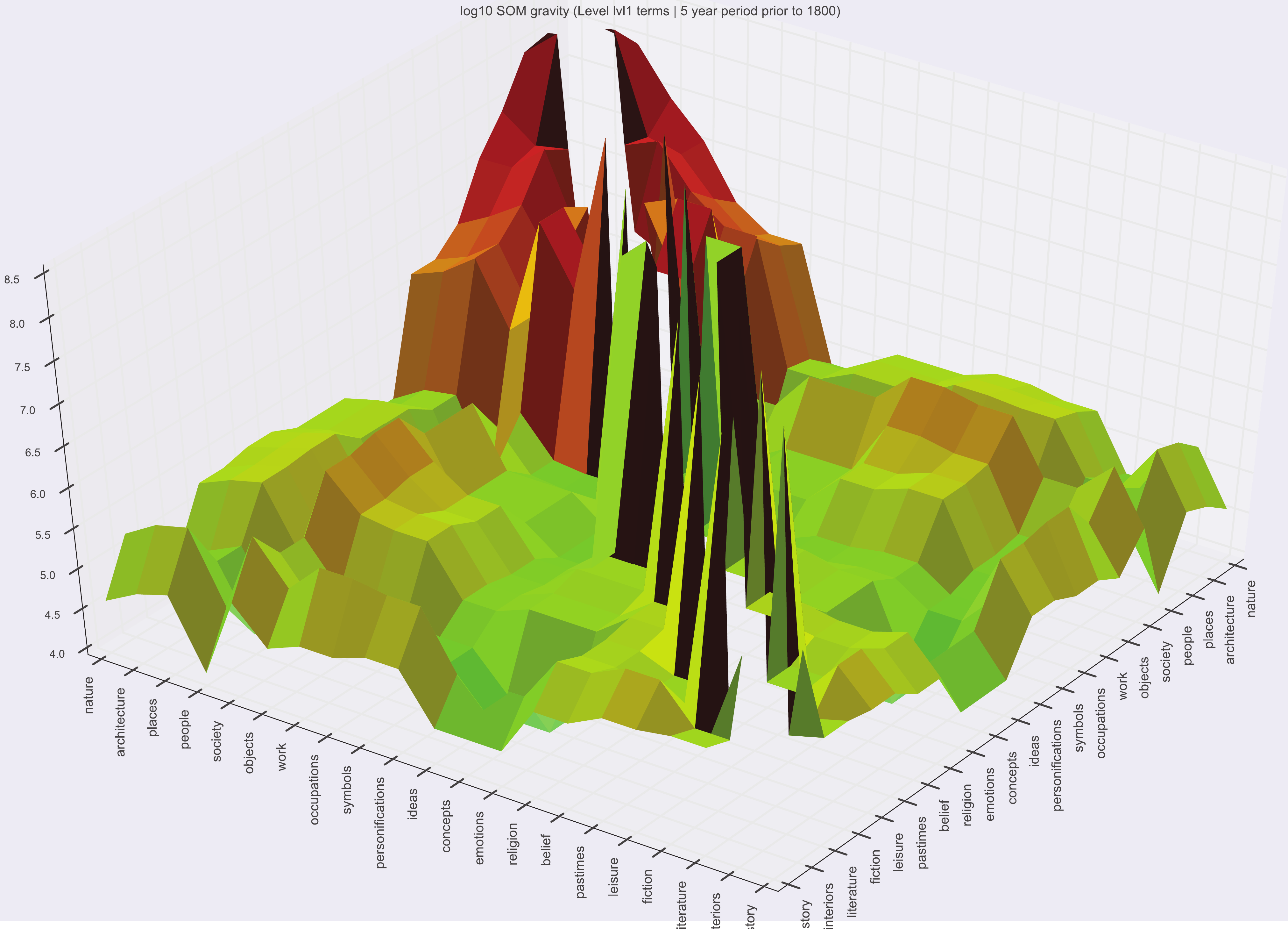}
        \end{subfigure}\hfill
        \begin{subfigure}[b]{0.20\textwidth}
                \centering
                \includegraphics[width=\linewidth]{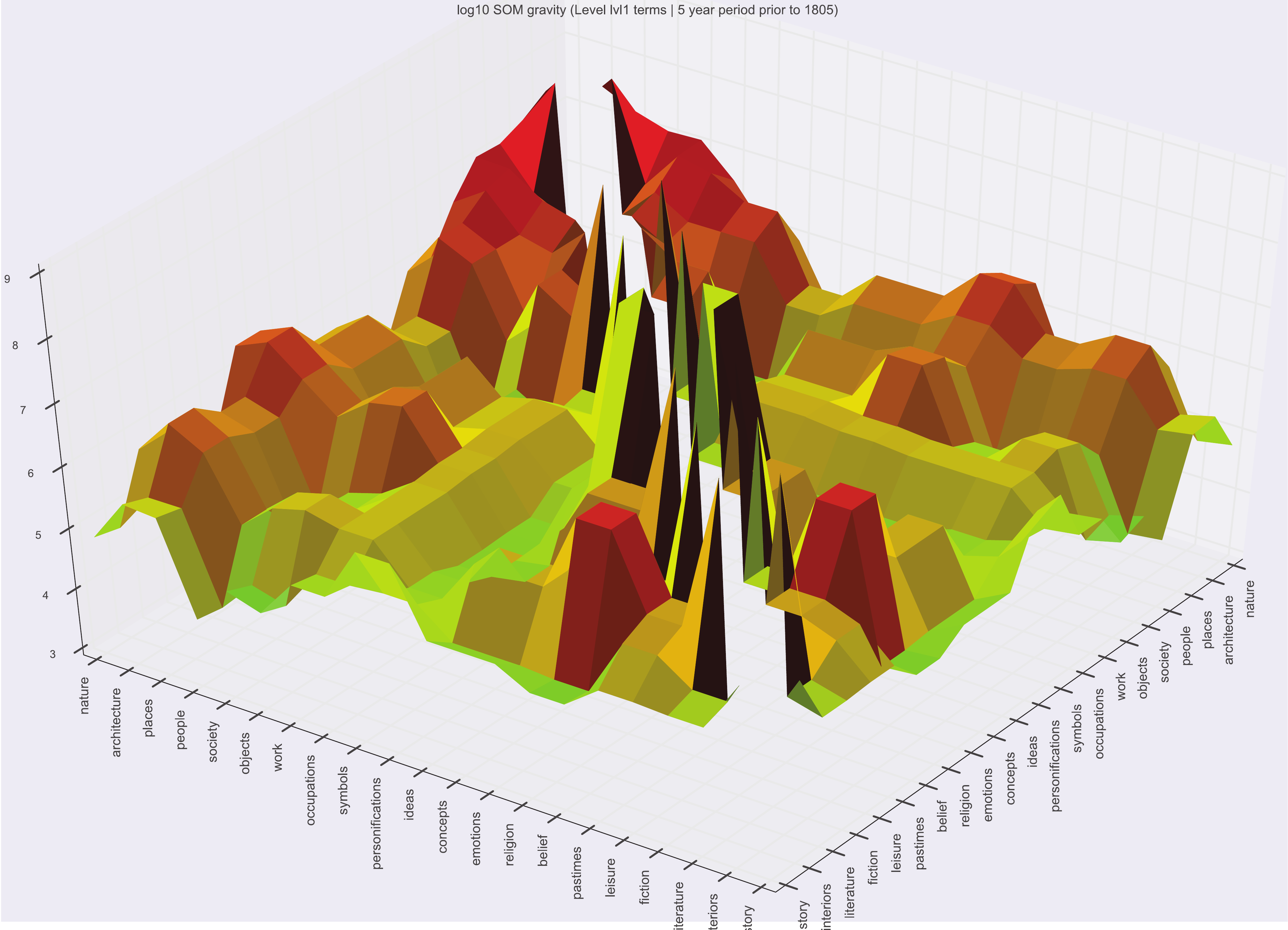}
        \end{subfigure}\hfill
        \begin{subfigure}[b]{0.20\textwidth}
                \centering
                \includegraphics[width=\linewidth]{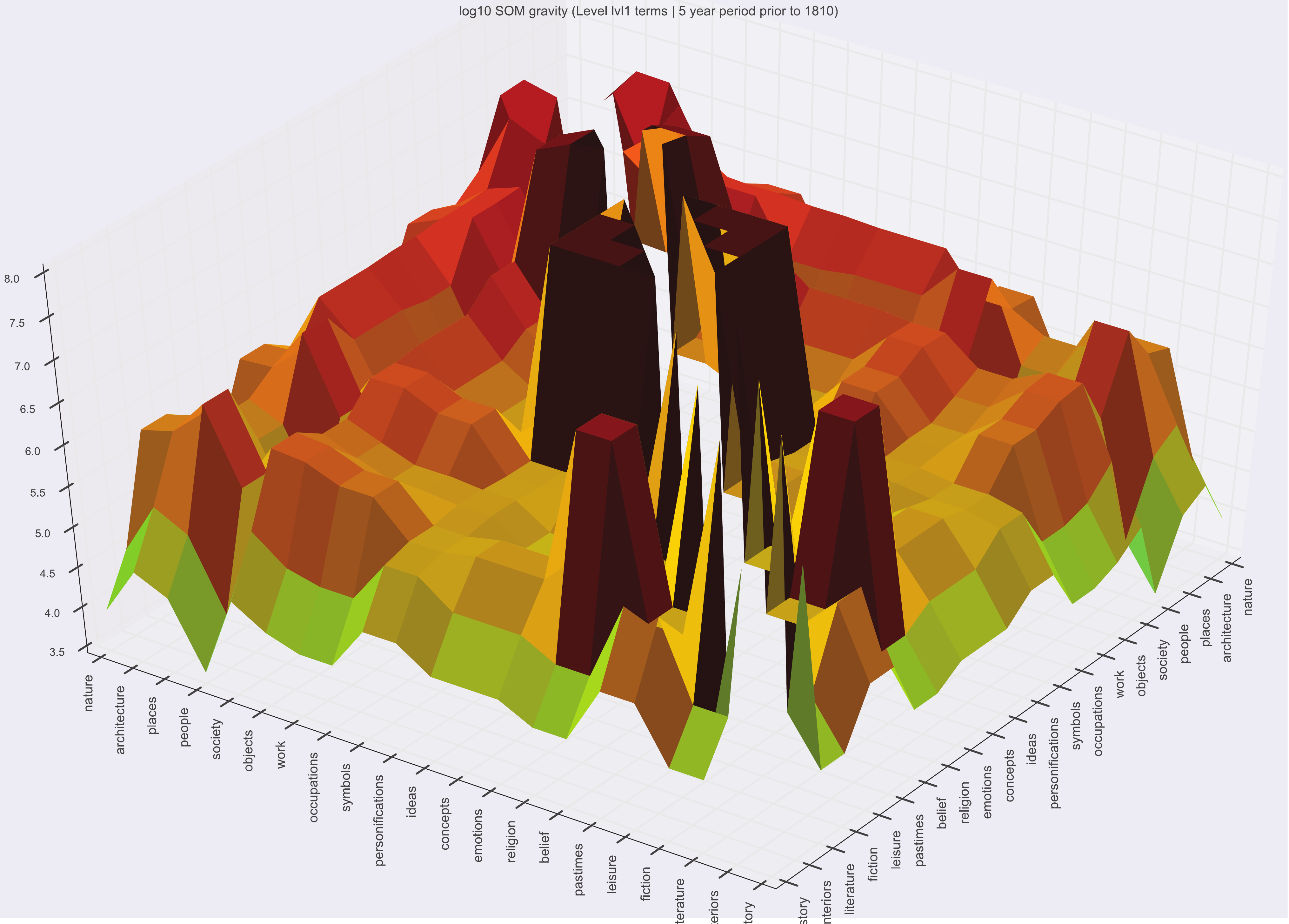}
        \end{subfigure}\hfill
        \begin{subfigure}[b]{0.20\textwidth}
                \centering
                \includegraphics[width=\linewidth]{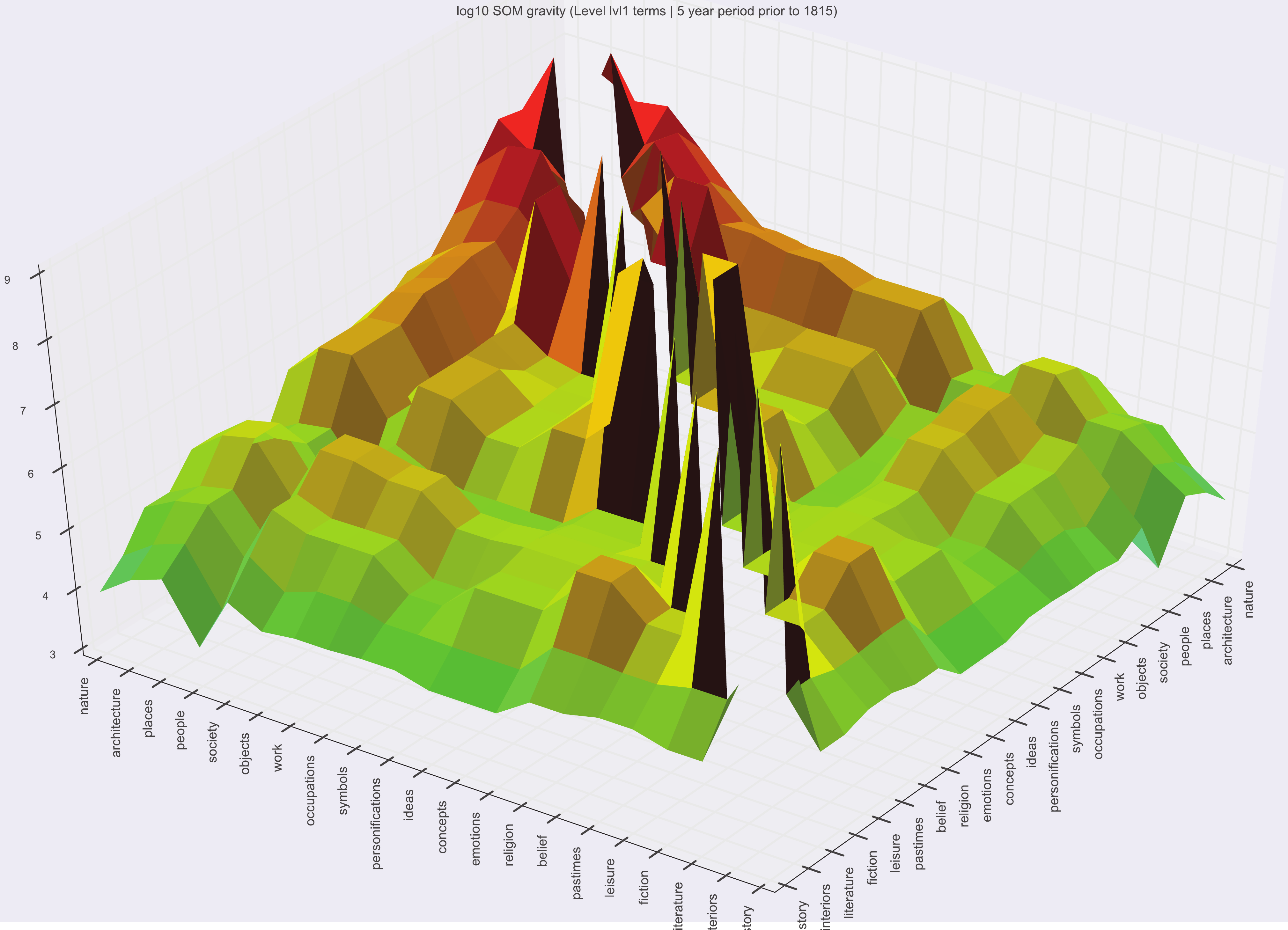}
        \end{subfigure}\hfill
        \begin{subfigure}[b]{0.20\textwidth}
                \centering
                \includegraphics[width=\linewidth]{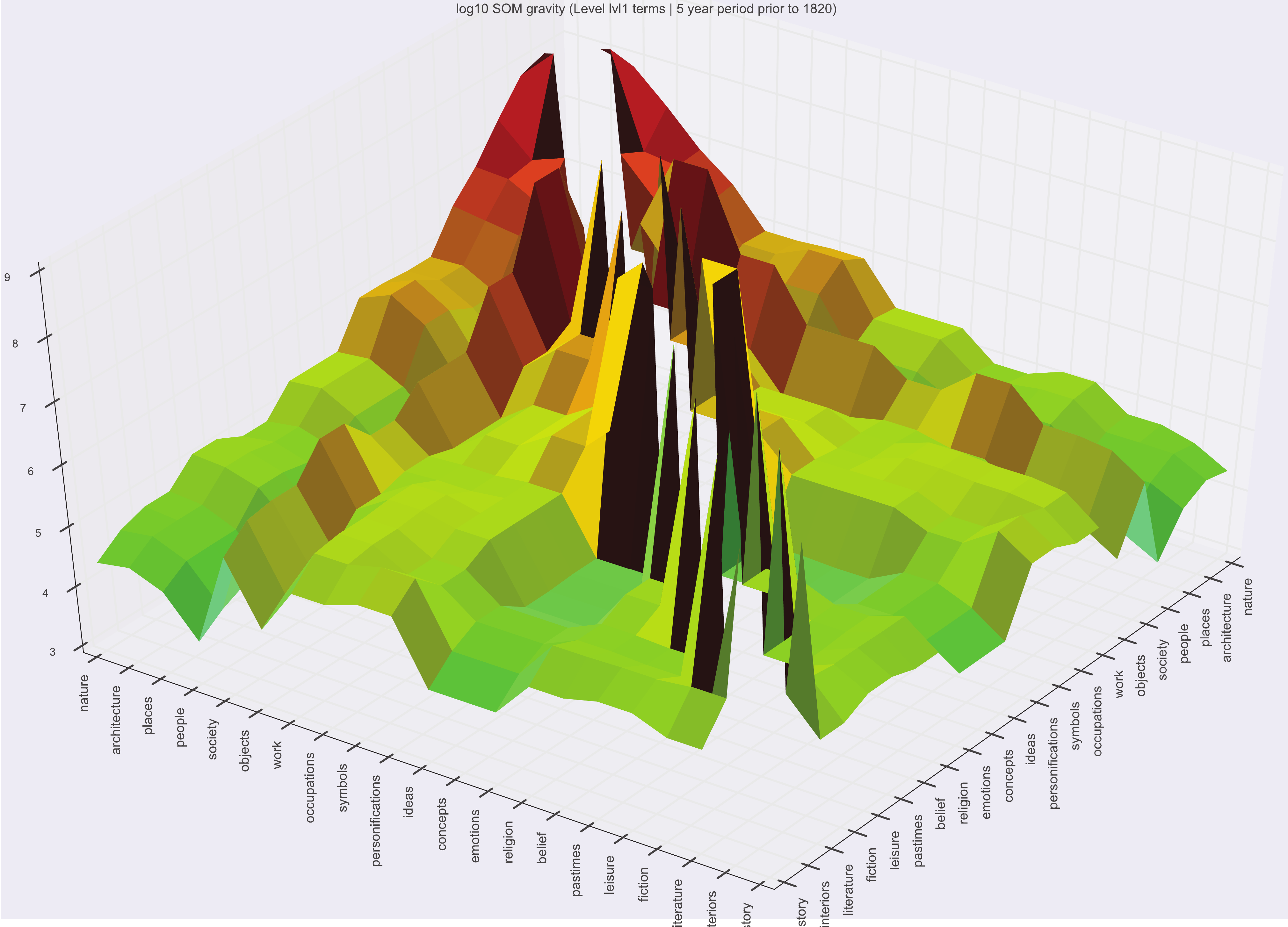}
        \end{subfigure}\hfill
				\\
        \begin{subfigure}[b]{0.20\textwidth}
                \centering
                \includegraphics[width=\linewidth]{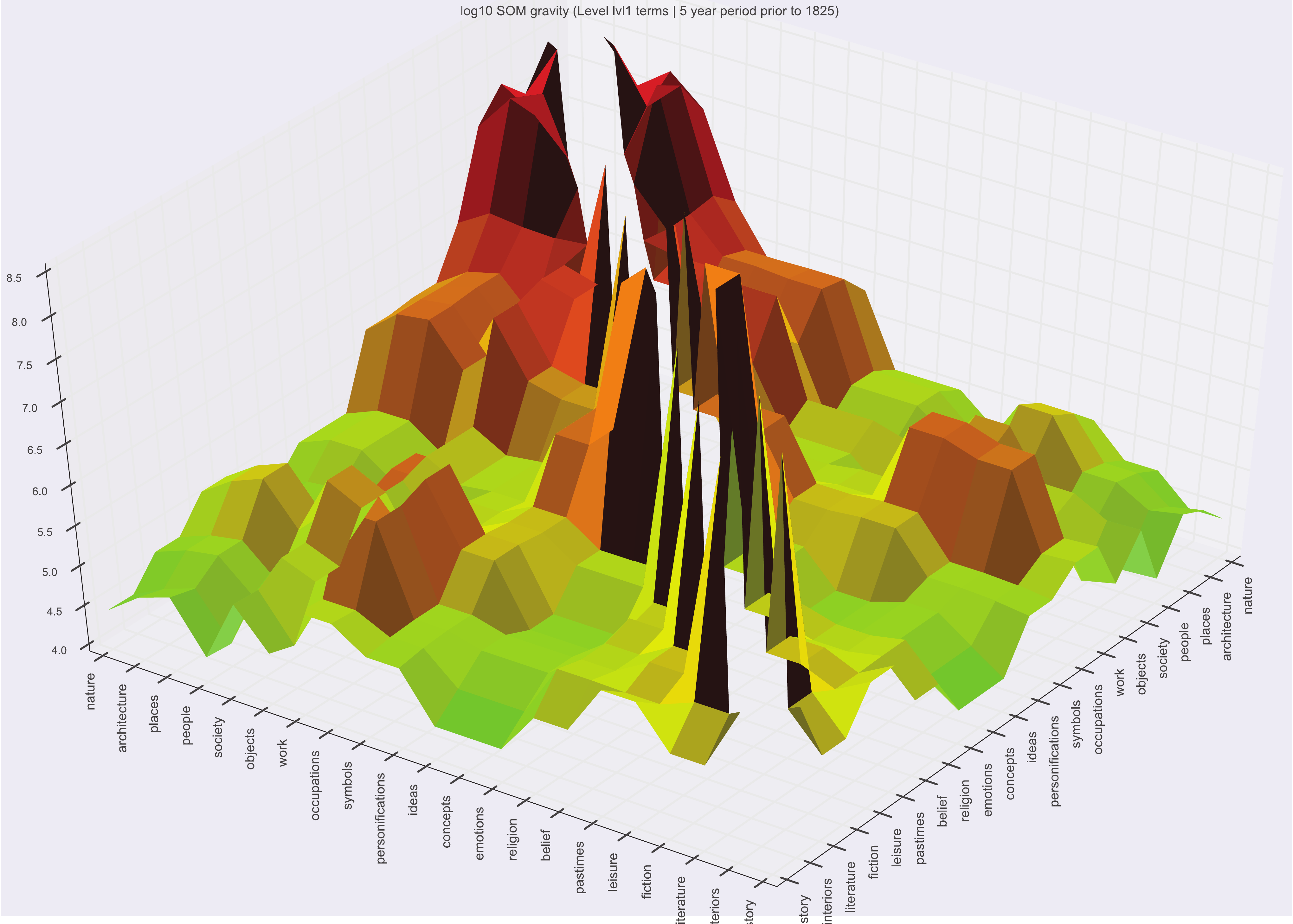}
        \end{subfigure}\hfill
        \begin{subfigure}[b]{0.20\textwidth}
                \centering
                \includegraphics[width=\linewidth]{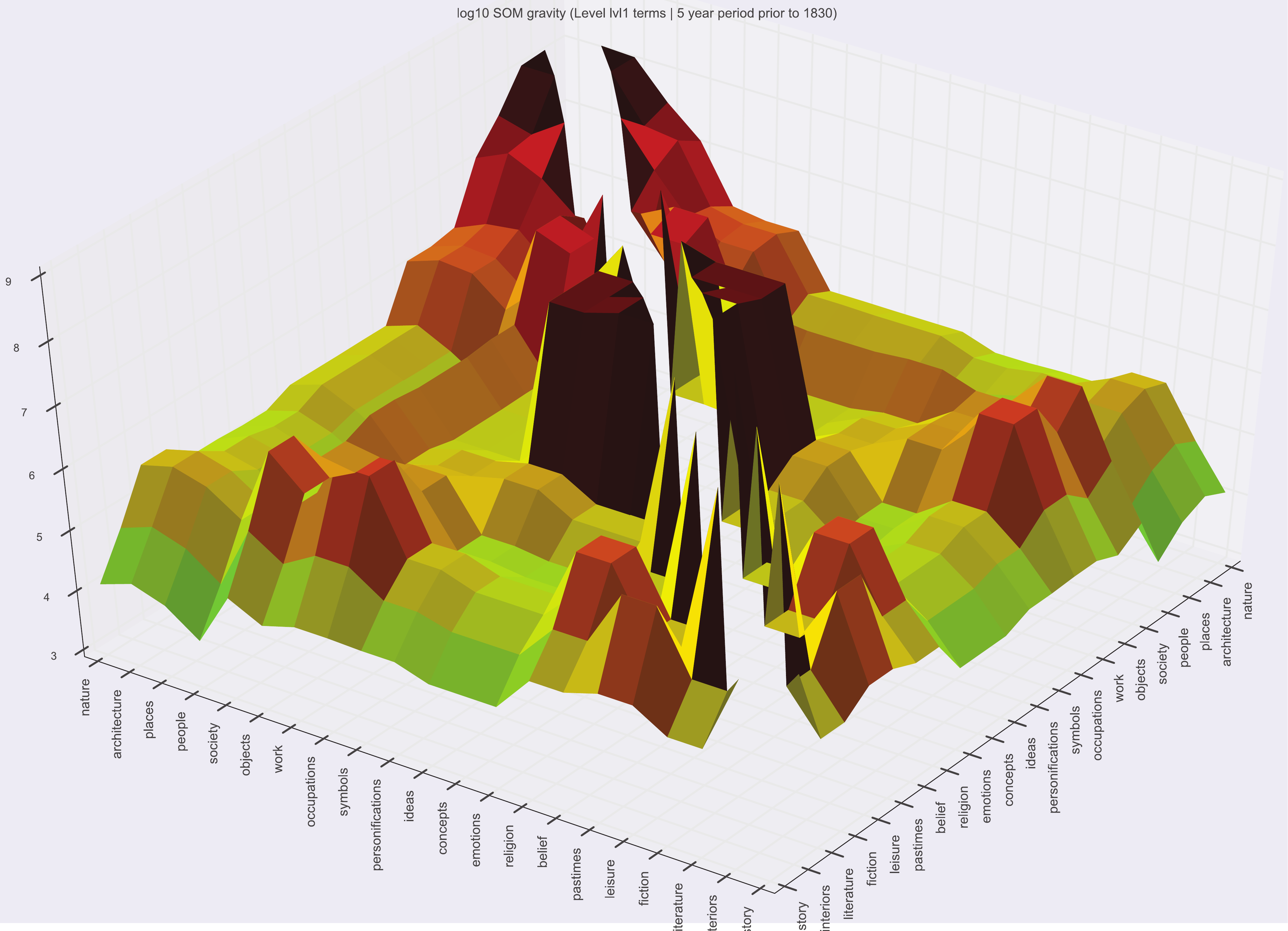}
        \end{subfigure}\hfill
        \begin{subfigure}[b]{0.20\textwidth}
                \centering
                \includegraphics[width=\linewidth]{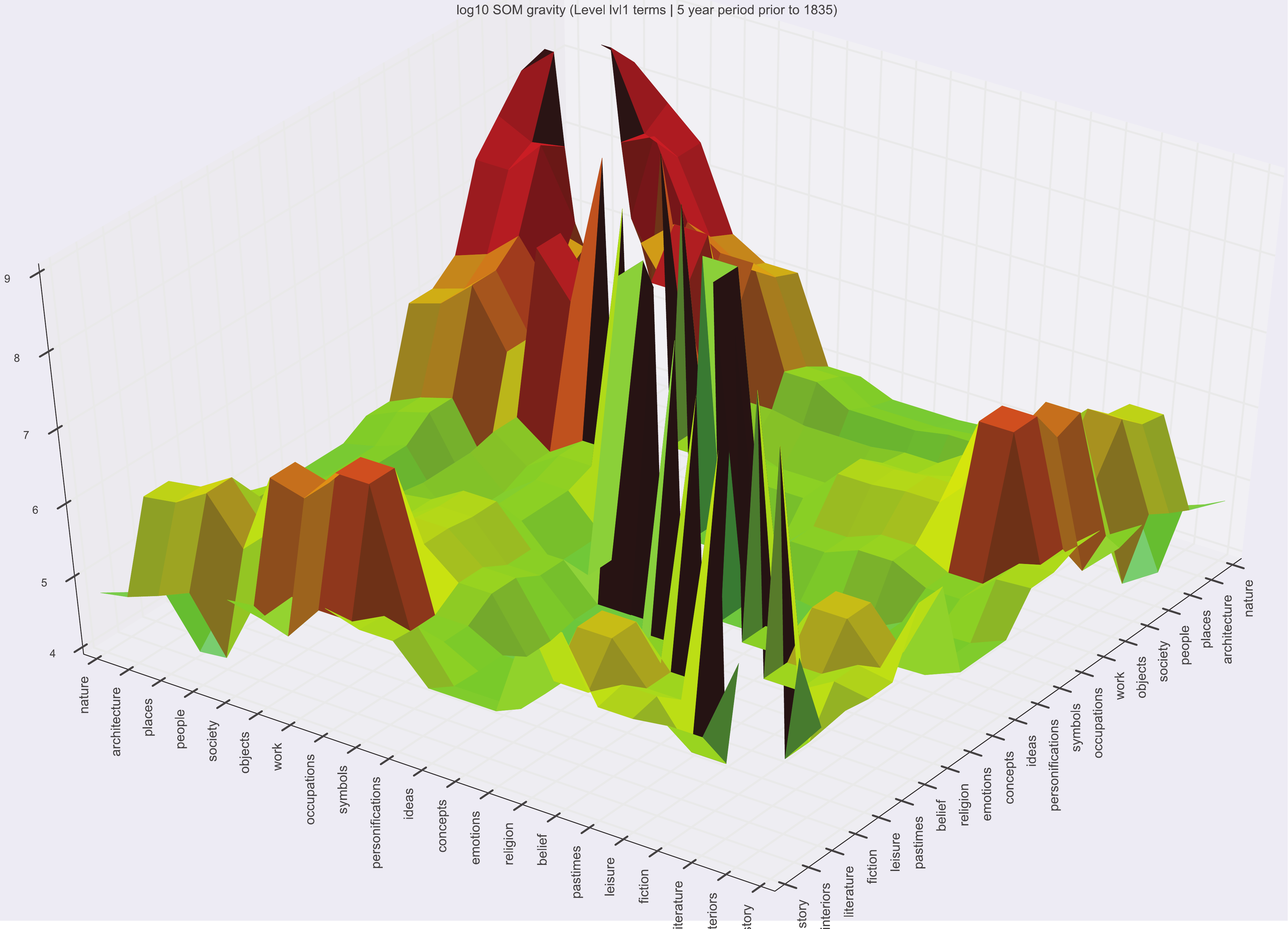}
        \end{subfigure}\hfill
        \begin{subfigure}[b]{0.20\textwidth}
                \centering
                \includegraphics[width=\linewidth]{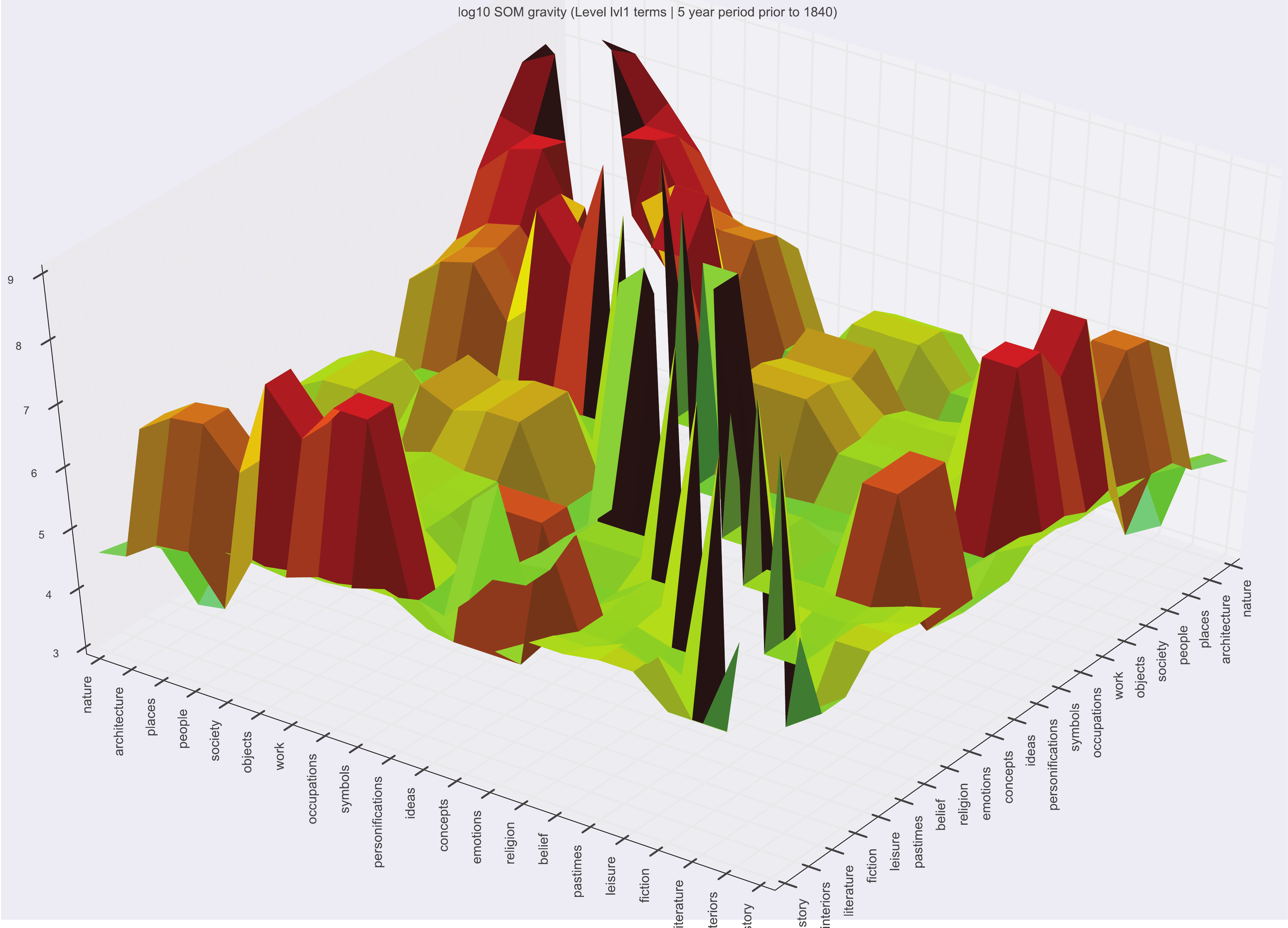}
        \end{subfigure}\hfill
        \begin{subfigure}[b]{0.20\textwidth}
                \centering
                \includegraphics[width=\linewidth]{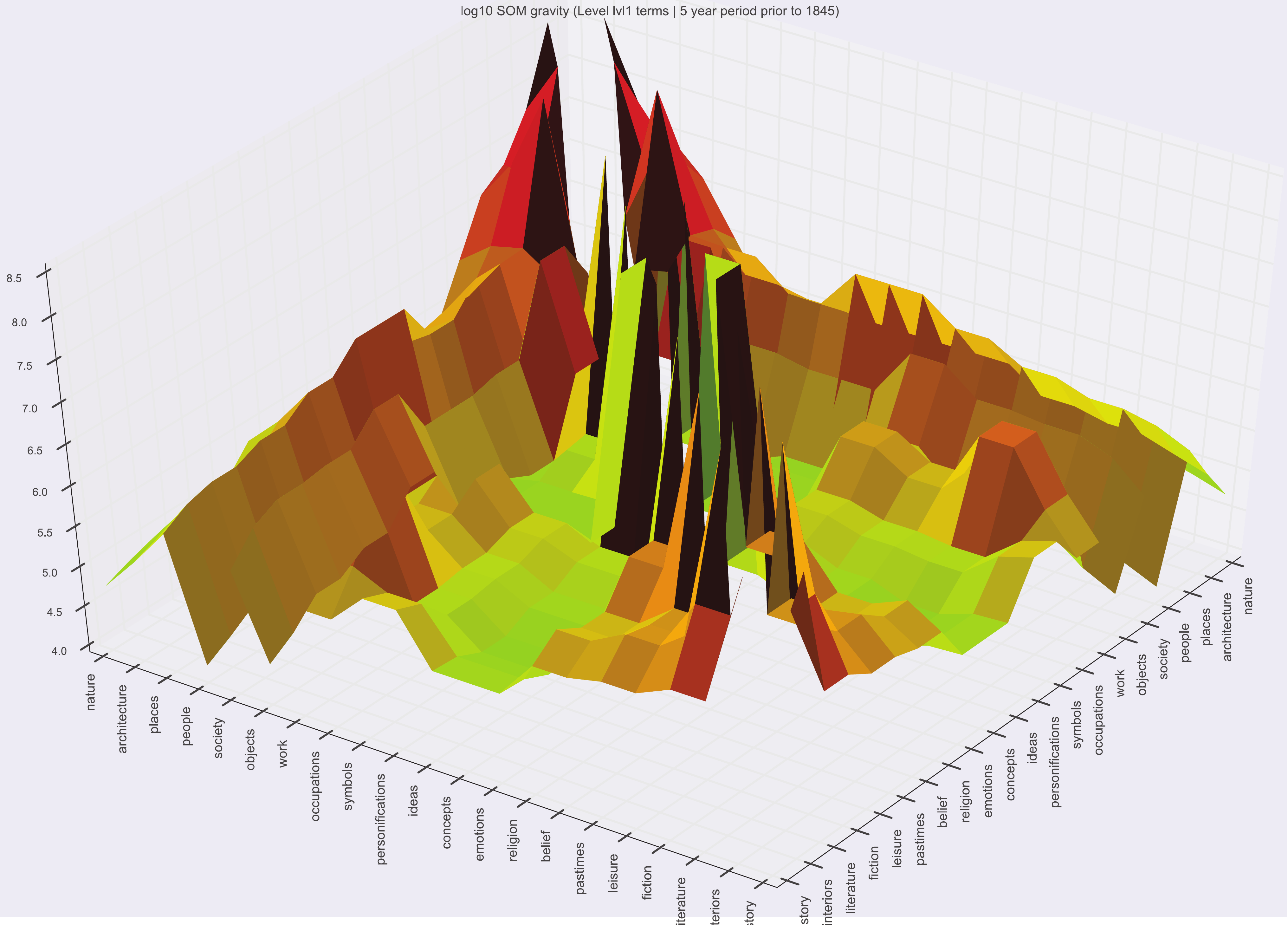}
        \end{subfigure}\hfill
				\\
				(b)\\
        \begin{subfigure}[b]{0.20\textwidth}
                \centering
                \includegraphics[width=\linewidth]{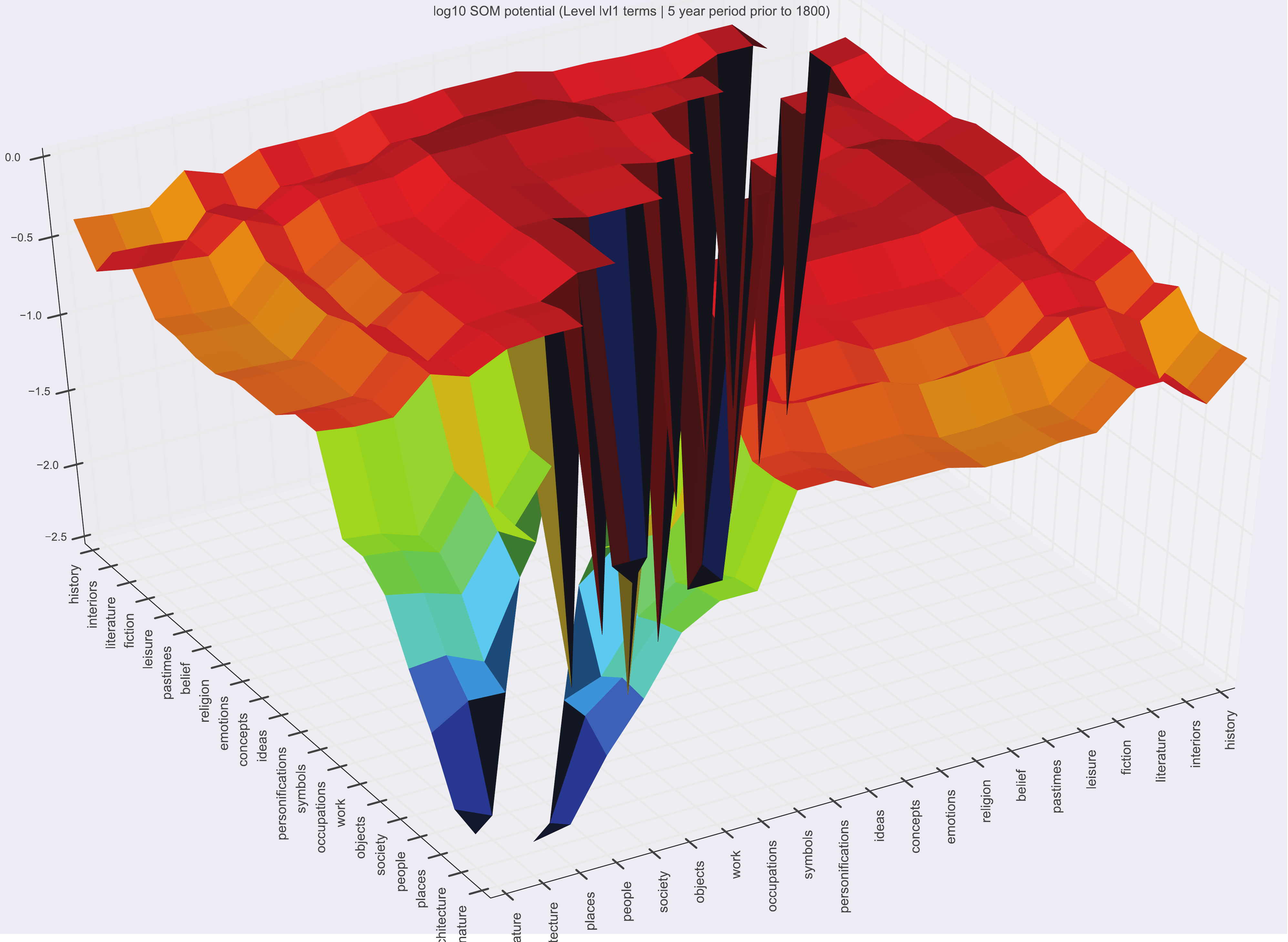}
        \end{subfigure}\hfill
        \begin{subfigure}[b]{0.20\textwidth}
                \centering
                \includegraphics[width=\linewidth]{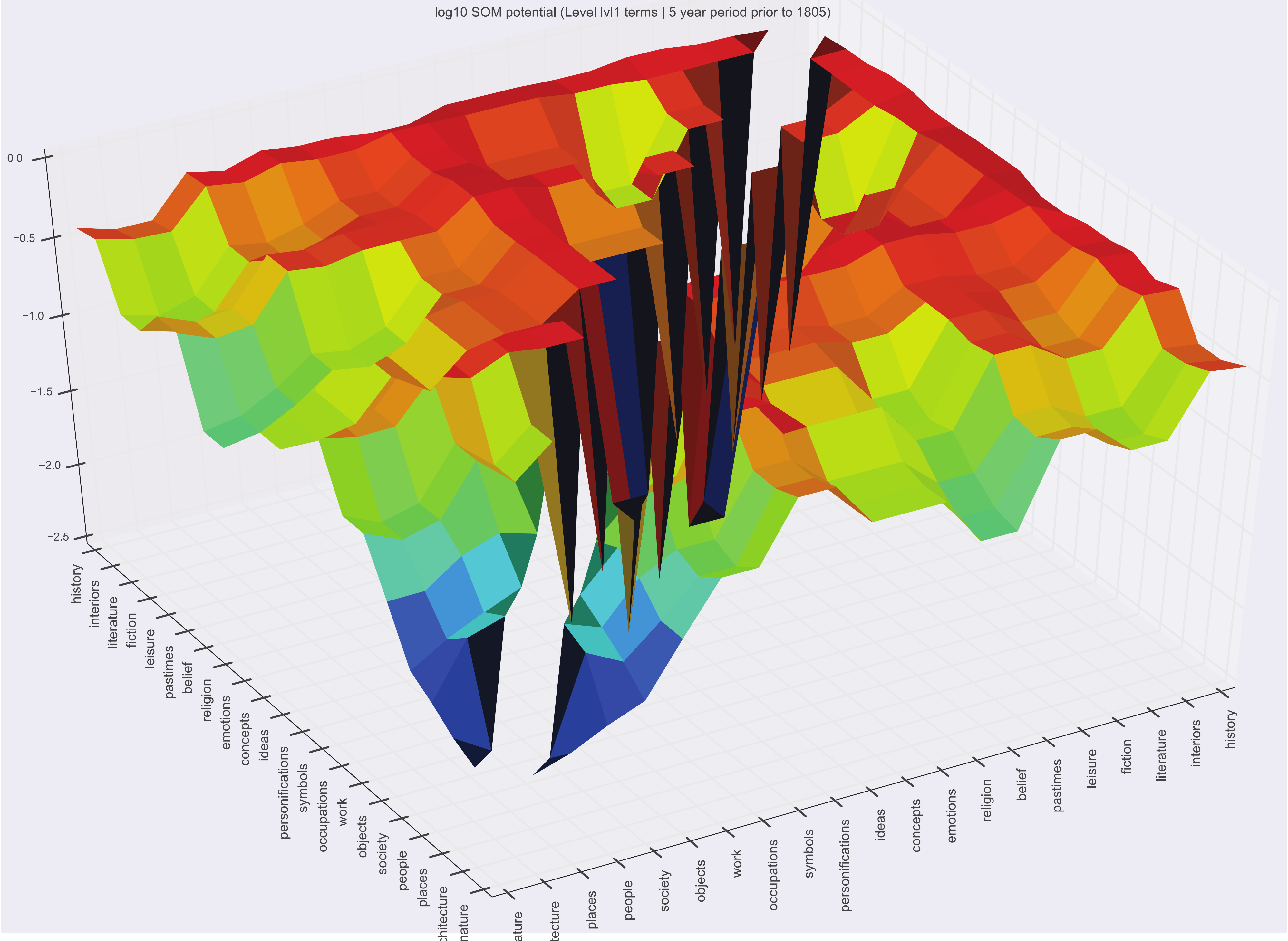}
        \end{subfigure}\hfill
        \begin{subfigure}[b]{0.20\textwidth}
                \centering
                \includegraphics[width=\linewidth]{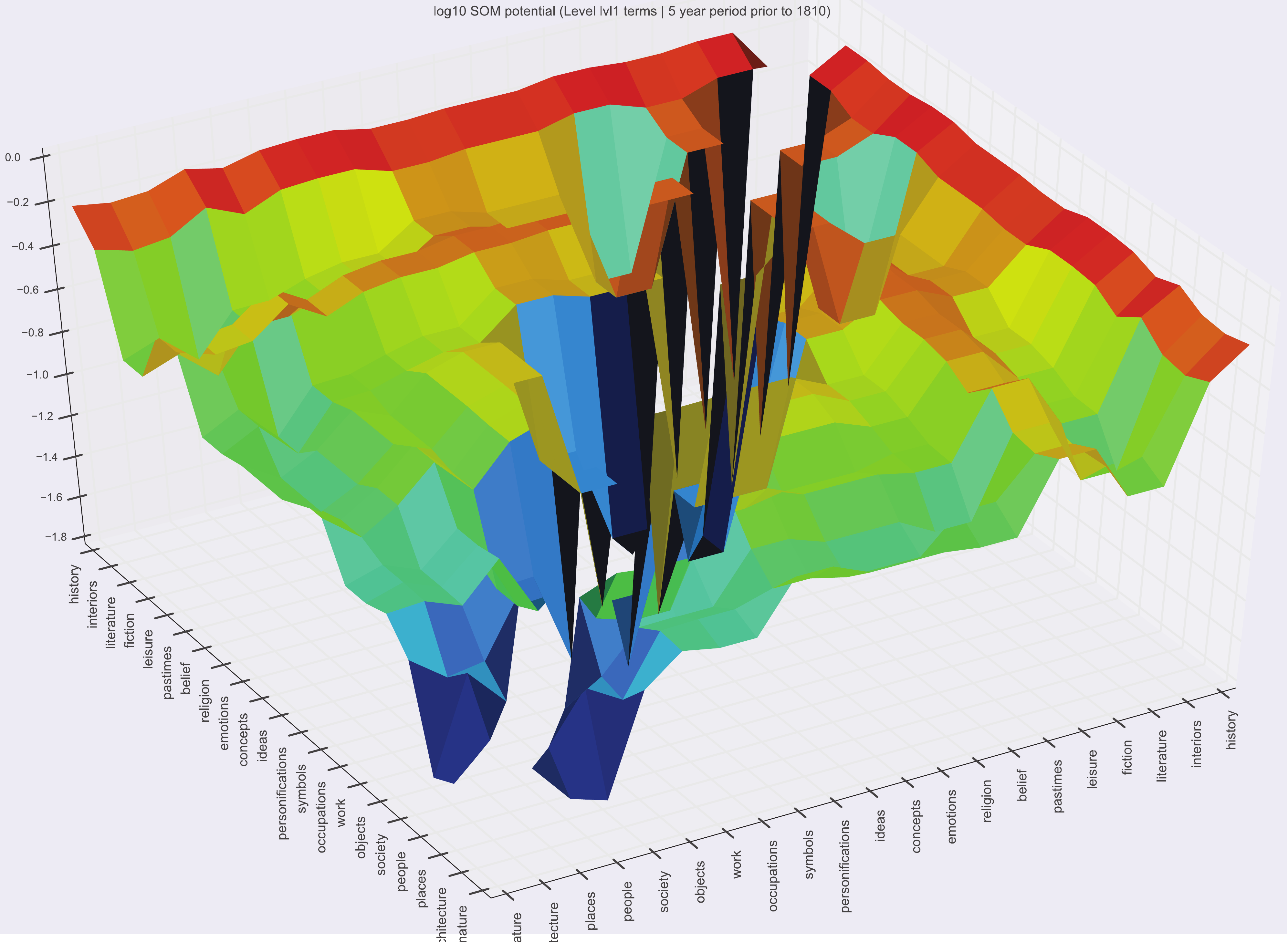}
        \end{subfigure}\hfill
        \begin{subfigure}[b]{0.20\textwidth}
                \centering
                \includegraphics[width=\linewidth]{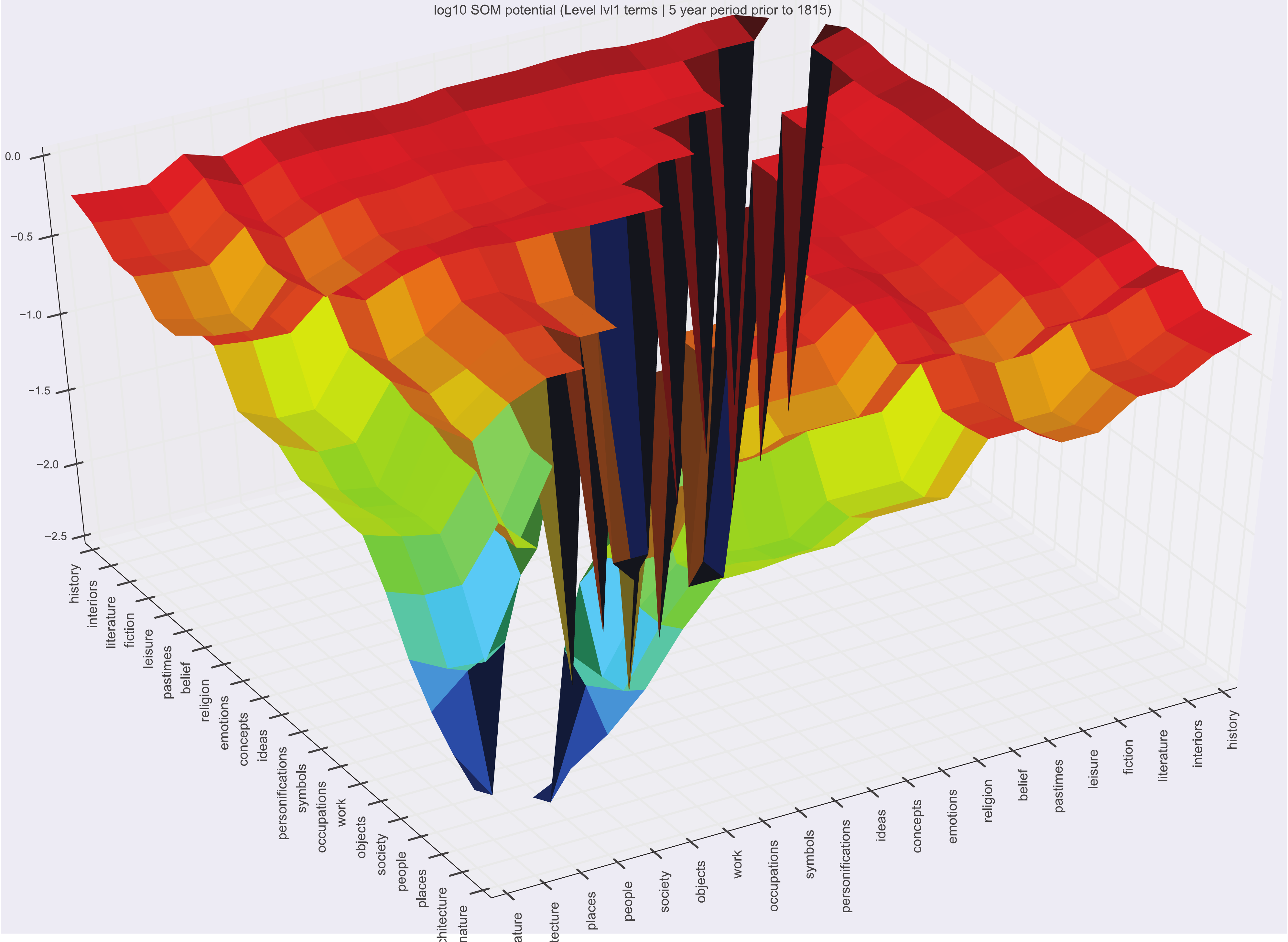}
        \end{subfigure}\hfill
        \begin{subfigure}[b]{0.20\textwidth}
                \centering
                \includegraphics[width=\linewidth]{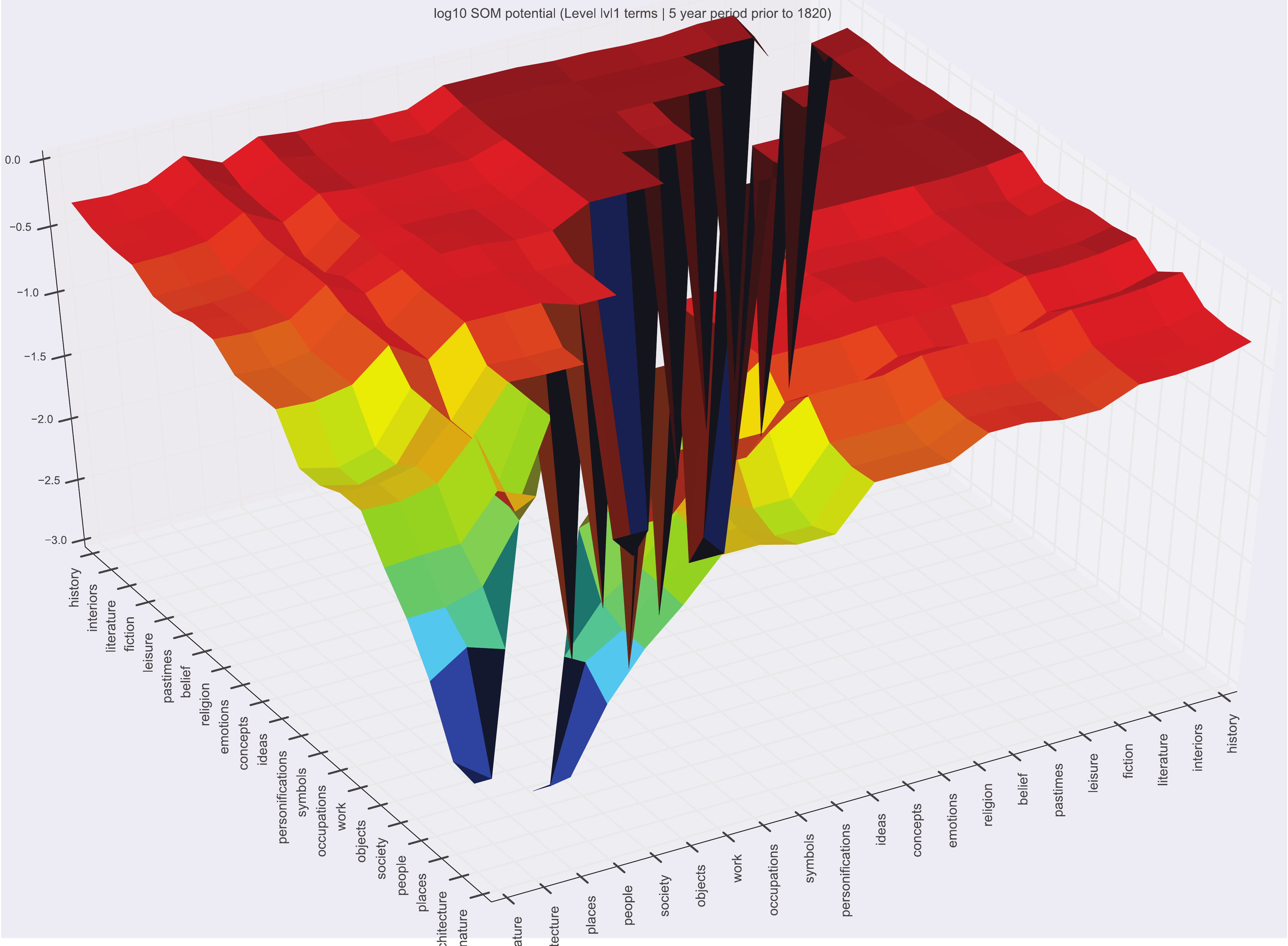}
        \end{subfigure}\hfill
				\\
        \begin{subfigure}[b]{0.20\textwidth}
                \centering
                \includegraphics[width=\linewidth]{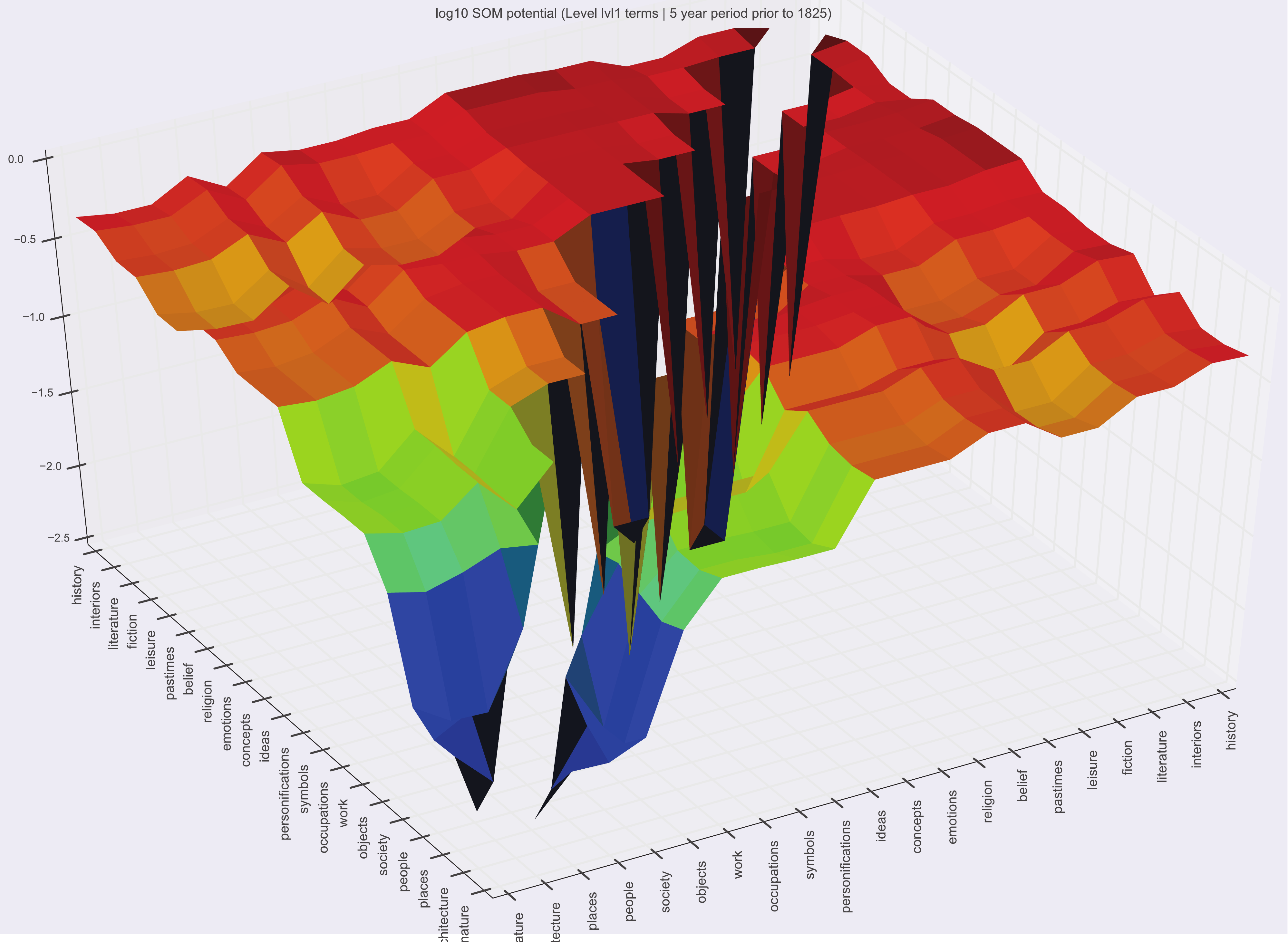}
        \end{subfigure}\hfill
        \begin{subfigure}[b]{0.20\textwidth}
                \centering
                \includegraphics[width=\linewidth]{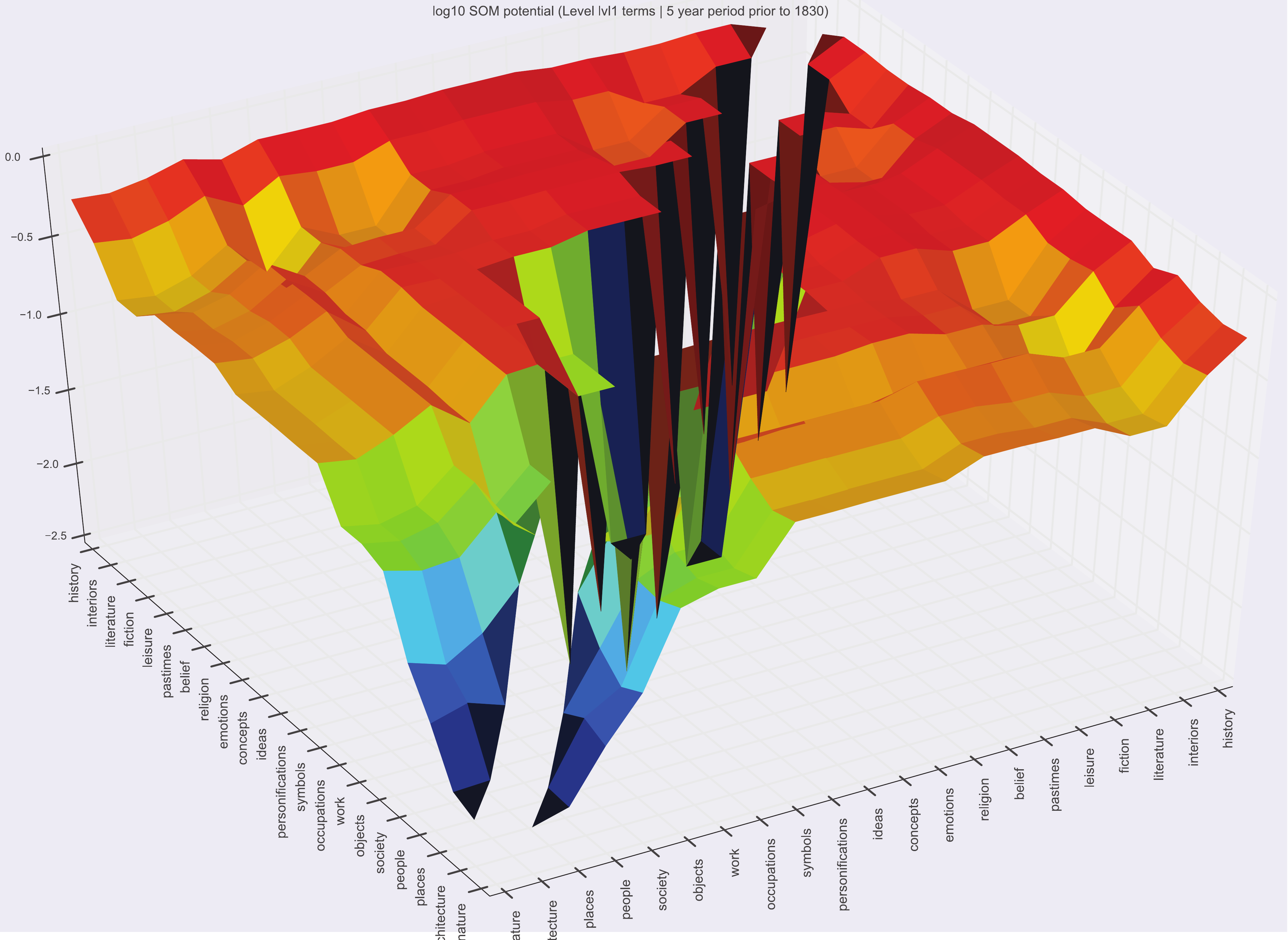}
        \end{subfigure}\hfill
        \begin{subfigure}[b]{0.20\textwidth}
                \centering
                \includegraphics[width=\linewidth]{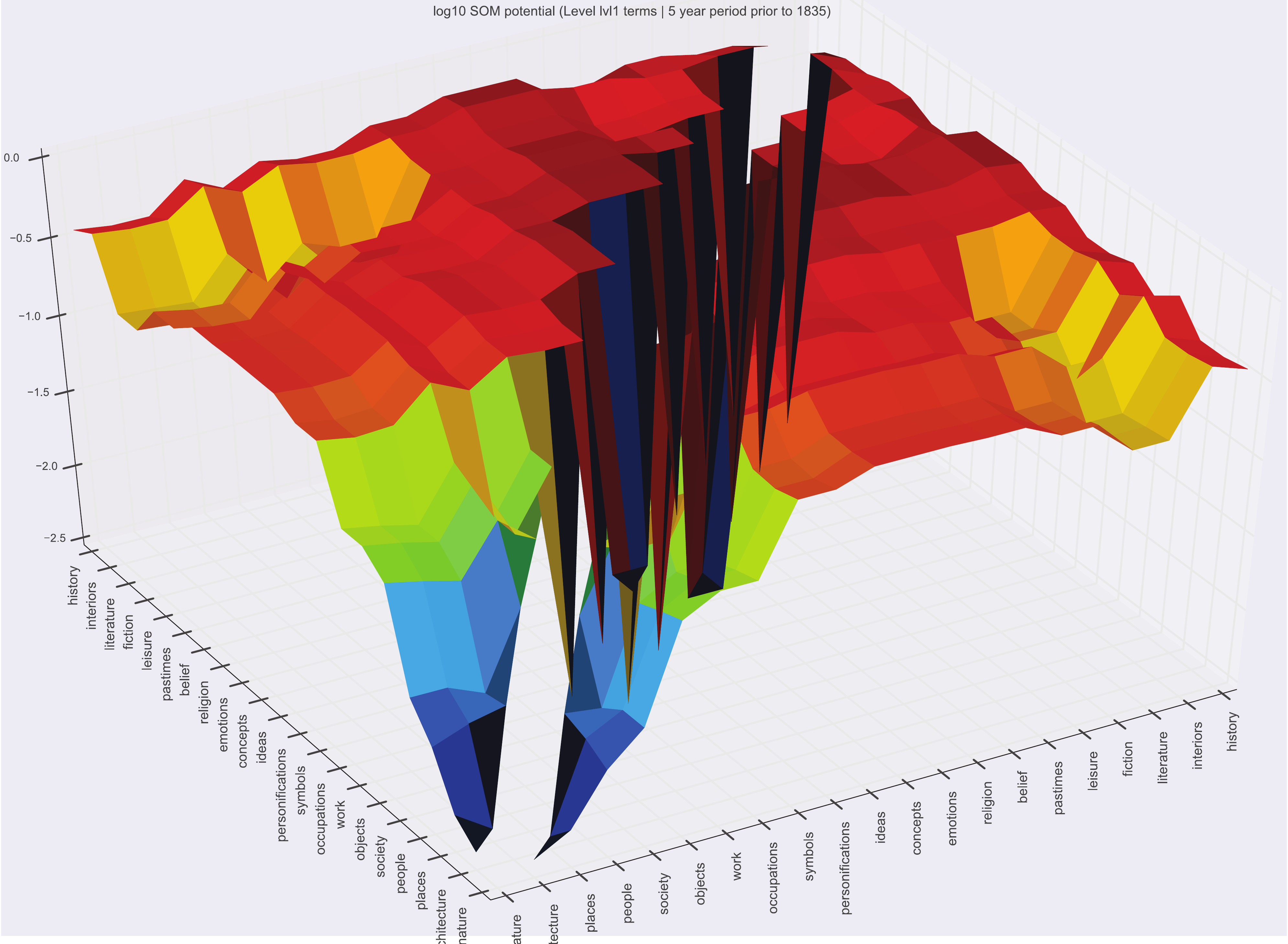}
        \end{subfigure}\hfill
        \begin{subfigure}[b]{0.20\textwidth}
                \centering
                \includegraphics[width=\linewidth]{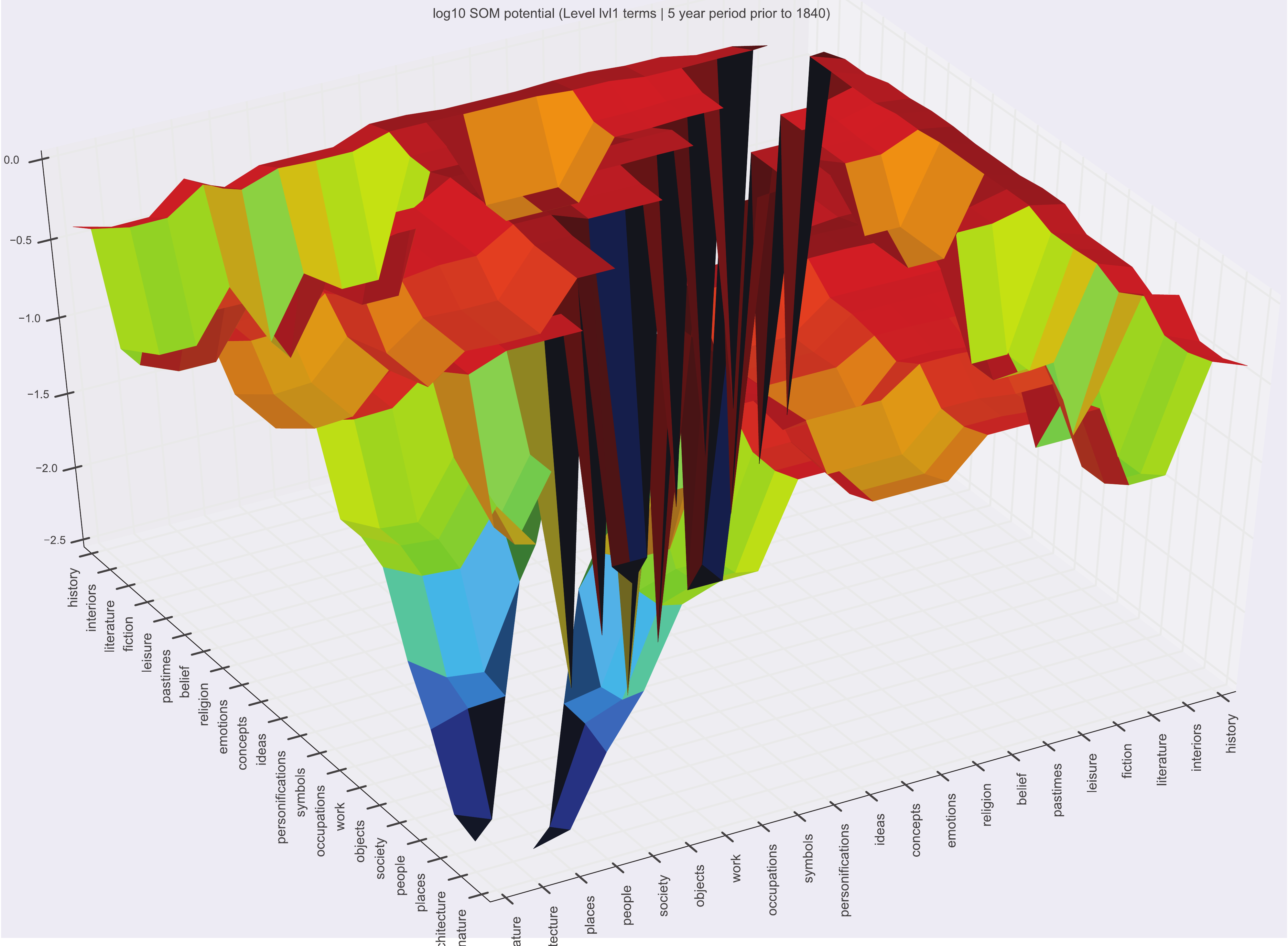}
        \end{subfigure}\hfill
        \begin{subfigure}[b]{0.20\textwidth}
                \centering
                \includegraphics[width=\linewidth]{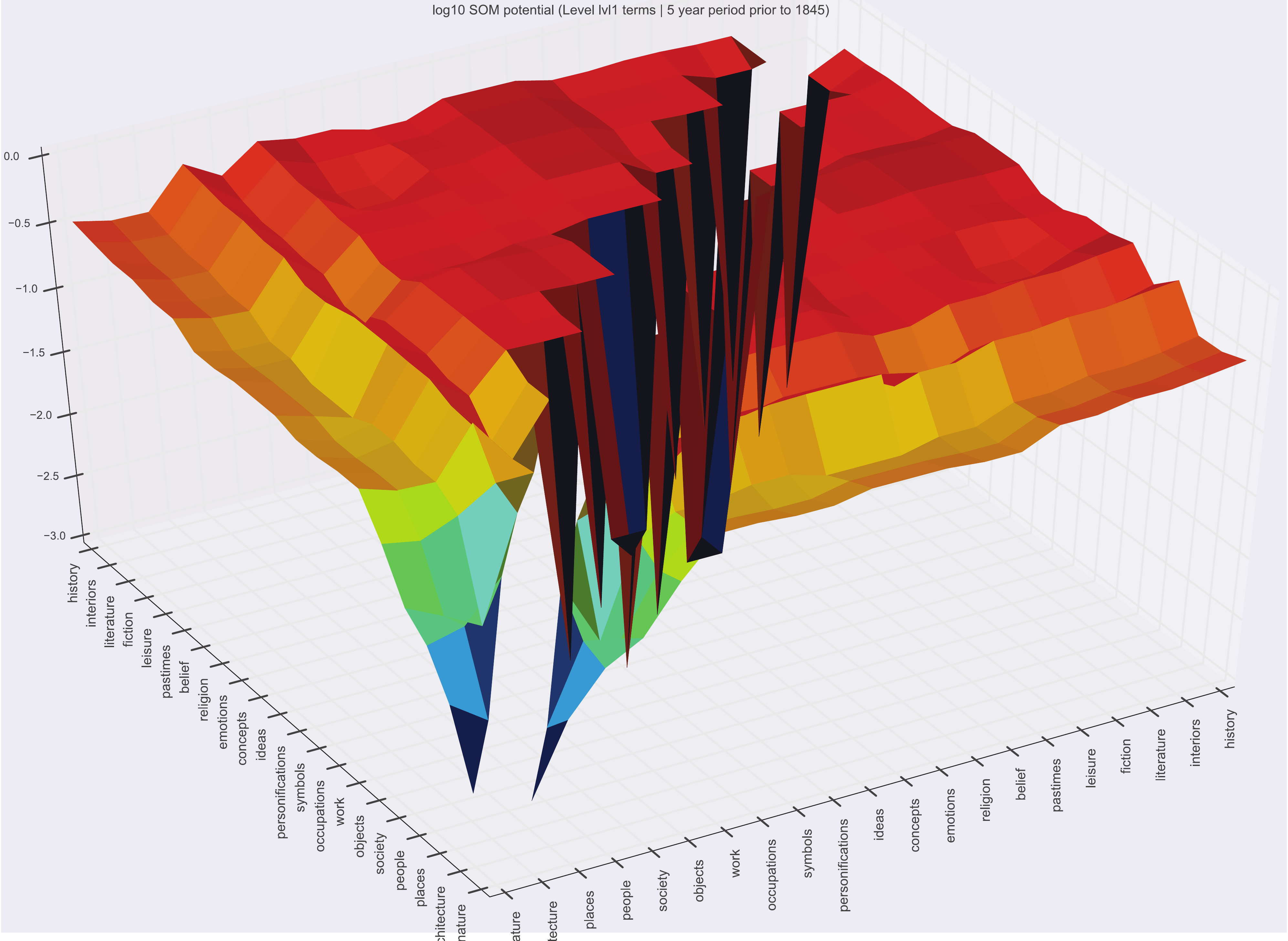}
        \end{subfigure}
        \caption{(a) Changes in the top [level 1] conceptual layer of the Tate indexing vocabulary in 1796-1845, sampled every 5 years, modeled on a gravitational field. Gravitational force is the negative gradient of the corresponding potential. (b) Respective changes in the underlying potential field. Extreme values indicate semantically related term pairs with high social status expressed by PageRank.}
				\label{Fig:2}
\end{figure*}

As we were left with the impression that in a statistically constructed vector field of term semantics drifts are the norm and not the exception, to account for such dynamics we computed a series of epoch-specific gravitational fields and their generating potential for a first overview. With BMU vector distances between term pairs and their PageRank values for `term mass', both types of surfaces expressed the interplay between semantic similarity and term importance in a social perspective (Figure \ref{Fig:2}).

\section{CONCLUSIONS AND FUTURE WORK}
In the above test, we resolved semantic drift detection, drift measurement, and partly resolved drift interpretation by the automatic evaluation of term cluster consistency. 
For the detection task, our detailed and thoroughly documented findings indicated that in an evolving collection, as could be expected from the idea of the dynamic library where vector space update results in displaced cluster centroids~\cite{salton1975dynamic}, drifts occur on a regular basis and become more frequent with increasing index term specificity. Apart from surveying the evolving semantic content structure, Somoclu also mapped the parallel evolution of classification tension structure, a precondition to future modeling and anomaly prediction.

Further we computed those evolving epoch-specific potential surfaces whose negative gradient was term similarity combined with term importance as an attractive force between feature or object pairs. This potential can be seen as the conceptual consequence of the semantic differential~\cite{osgood1957mmu}, a forerunner to modern latent semantic methods. This semantic potential, in turn, suggests that physics as a metaphor is useful because it yields new, helpful concepts to model the dynamics of meaning, itself important for knowledge organization and knowledge management.

Our effort belongs to the field of \textit{social mechanics}, a 21st century repercussion of ideas dating back as far as 1769 when American political theorist James Madison (1751-1836), the so-called `father of the constitution' and the United States' fourth president, was said to be studying a primitive form of it at Princeton. After him and over the centuries to come, prominent thinkers often tried to understand society's workings e.g. by means of thermodynamics or mechanics. In our implementation, social mechanics is a variant of classical mechanics because the concept of mass we apply to features in general and index terms in particular, is a relative (evolving) one, depending on language use as its social context and implemented by the distributional hypothesis. 

By doing so, the `meaning as change' paradigm receives experimental support inasmuch as `term mass' corresponds to work investment during update, with the reconfiguration of semantic spaces and fields being proportional to it. In order to explore the semantic potential, to connect measures of semantic relatedness with centrality values such as PageRank for `term mass' will be subject to future research, with substantial input expected e.g. from~\cite{petitot2004morphogenesis} or~\cite{cooper1999linguistic}. 

\section{ACKNOWLEDGMENTS}
This research received funding by the European Commission Seventh Framework Programme under Grant Agreement Number FP7-601138 PERICLES. S\'andor Dar\'anyi is grateful to Emma Tonkin (University of Bristol) for early discussions on the subject.

\balancecolumns
\end{document}